%% file: main.tex
\documentclass{article}

\usepackage{preprint,times}
\usepackage{authblk} 

\input{math_commands.tex}

\usepackage{graphicx}
\usepackage{wrapfig}
\usepackage{booktabs}
\usepackage{tabularx}
\usepackage{array}
\usepackage{multirow}
\usepackage[normalem]{ulem}

\usepackage{placeins}
\usepackage{tikz}
\usetikzlibrary{arrows.meta,positioning}
\usepackage{url}
\usepackage{pifont}
\usepackage{colortbl}
\usepackage{float}
\usepackage[table]{xcolor}
\usepackage{needspace}
\usepackage{amssymb}
\usepackage[caption=false,font=footnotesize]{subfig}

\definecolor{evalaction}{RGB}{235,244,249}
\definecolor{evalgt}{RGB}{244,239,248}

\definecolor{nvgreen}{RGB}{136,204,2}

\usepackage[colorlinks=true,
            linkcolor=nvgreen,      
            citecolor=nvgreen,  
            urlcolor=nvgreen]{hyperref}

\definecolor{parcontent}{HTML}{5B6770}
\definecolor{parcontenttint}{HTML}{F0F2F4}
\definecolor{parreference}{HTML}{0F8B8D}
\definecolor{parreferencetint}{HTML}{E6F4F4}
\definecolor{gtxt}{HTML}{5B6470}
\definecolor{dsnormal}{HTML}{EAF3EF}\definecolor{dsood}{HTML}{F8F1E5}\definecolor{dssafety}{HTML}{EAF1F6}
\definecolor{entitygreen}{HTML}{0B5D1E}

\newcommand{\ours}{\textbf{\method{}}}
\newcommand{\mrow}{\hspace{2pt}}
\newcommand{\hz}[1]{\textcolor{gtxt}{\scriptsize #1}}
\newcommand{\gin}[1]{#1}

\graphicspath{{fig/}}
\newcommand{\method}{GroundAct}

\newcolumntype{Y}{>{\raggedright\arraybackslash}X}

\input{fig/eval_mode_icons.tex}

\title{\texorpdfstring{\raggedright\raisebox{-0.05ex}{\includegraphics[height=1.5ex]{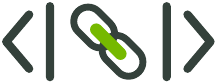}}\hspace{0.35em}}{}Grounding What Shapes the Plan: \mbox{Rethinking} Groundedness for Physical Intelligence in Autonomous Driving}

\renewcommand\Affilfont{\normalfont\normalsize}

\makeatletter
\renewcommand\AB@affilsepx{\hspace{1.5em}\protect\Affilfont}
\renewcommand\AB@affilnote[1]{}
\makeatother

\author[1,2]{Minkyoung Cho\textsuperscript{$\dagger$}}
\author[1,3]{Zewei Zhou\textsuperscript{$\dagger$}}
\author[1]{Wenhao Ding}
\author[1]{Shuhan Tan}
\author[1]{Boyi Li}
\author[1]{\\Yuxiao Chen}
\author[1]{Yan Wang}
\author[1]{Zheng Lian}
\author[1]{Min-Hung Chen}
\author[1,4]{Chaowei Xiao}
\author[5]{\\Z. Morley Mao}
\author[1]{Boris Ivanovic}
\author[1,6]{Marco Pavone}
\author[1]{Yulong Cao}

\affil[3]{NVIDIA\textsuperscript{1}}
\affil[1]{UMich\textsuperscript{2}}
\affil[2]{UCLA\textsuperscript{3}}
\affil[4]{JHU\textsuperscript{4}}
\affil[5]{USC\textsuperscript{5}}
\affil[6]{Stanford\textsuperscript{6}}

\makeatletter
\vspace{2pt}
\g@addto@macro\AB@affillist{\\[0.4em] \url{https://groundact.github.io/}}
\makeatother

\newcommand\blfootnote[1]{%
  \begingroup
    \renewcommand\thefootnote{}\footnote{#1}%
    \addtocounter{footnote}{-1}%
  \endgroup
}

\begin{document}

\maketitle
\blfootnote{\textsuperscript{$\dagger$}Work done during an internship at NVIDIA.\quad Correspondence: \href{mailto:minkycho@umich.edu}{\nolinkurl{minkycho@umich.edu}}.}

\vspace{-12pt}
\begin{abstract}
\input{sec/0_abstract}

\end{abstract}

\input{sec/1_intro}
\input{sec/2_related_work}

\input{sec/3_method}

\input{sec/4_evaluation}
\input{sec/5_conclusion}

\bibliography{main}
\bibliographystyle{iclr2027_conference}

\input{sec/X_suppl}
\end{document}

%% file: math_commands.tex
\usepackage{amsmath,amsfonts,bm}

\def\eqref#1{equation~\ref{#1}}
\def\Eqref#1{Equation~\ref{#1}}

\def\1{\bm{1}}

\DeclareMathAlphabet{\mathsfit}{\encodingdefault}{\sfdefault}{m}{sl}
\SetMathAlphabet{\mathsfit}{bold}{\encodingdefault}{\sfdefault}{bx}{n}



%% file: fig/eval_mode_icons.tex
\ifdefined\evalmodeD\else
\definecolor{evalmodeblue}{HTML}{266FA8}
\definecolor{evalmodepurple}{HTML}{8651A3}
\DeclareRobustCommand{\evalmodeicon}[2]{%
  \begingroup
  \normalfont
  \tikz[baseline=-0.32em,line width=0.055em,line cap=round]{%
    \path[use as bounding box] (-0.60em,-0.60em) rectangle (0.60em,0.60em);
    \draw[draw=#2,fill=#2!10!white] (0,0) circle[radius=0.54em];
    \node[inner sep=0pt,outer sep=0pt,text=#2,
      font=\sffamily\bfseries\fontsize{0.80em}{0.80em}\selectfont]
      at (0,0) {#1};
  }%
  \endgroup
}
\DeclareRobustCommand{\evalmodeD}{\evalmodeicon{D}{evalmodeblue}}
\DeclareRobustCommand{\evalmodeC}{\evalmodeicon{C}{evalmodepurple}}
\fi

%% file: sec/0_abstract.tex

Driving models increasingly ground reasoning in causal relations,
spatial structure, perceptual evidence, and predicted futures. These advances make
reasoning more faithful to the driving scene, but leave a fundamental question
unresolved: \emph{what should groundedness mean when the model ultimately outputs an action?}
Correctly grounded reasoning does not, by itself, ensure desirable driving outcomes.
We introduce \textbf{\method{}}, which starts from a simple premise: driving unfolds
through physical entities and their interactions. Entities therefore become the
unit of grounding; a lightweight reference token keeps each selected entity's
continuous state addressable through symbolic reasoning; and only the referenced
entities' interactions with the evolving proposal correct the plan. The result is
an explicit path from what reasoning grounds to what the plan does, which we call
\emph{grounded planning}.
To assess its practical value, we evaluate GroundAct in both open- and closed-loop settings. 
GroundAct shows strong open-loop planning across normal, out-of-distribution, 
and safety-critical scenarios, with closed-loop results extending this evidence 
to driving in simulation.

%% file: sec/1_intro.tex

\section{Introduction}
\label{sec:introduction}

\vspace{-4pt}
Autonomous-driving models increasingly incorporate reasoning for its potential to support generalization across diverse and long-tail scenarios, grounding it in causal relationships, spatial structure, perceptual evidence, and predicted future states~\citep{futuresightdrive,dynvla,cf-vla,lcdrive}. Structured reasoning traces further connect grounded observations and predictions to high-level intentions for how the vehicle should behave~\citep{alpamayo}.
Collectively, these advances improve what a driving model can explain about the scene,
why a decision may be appropriate, and what may happen next.

\begin{figure}[t]
  \centering
  \begingroup
  \newlength{\figonepanelheight}
  \setlength{\figonepanelheight}{0.24\textwidth}
  \begin{minipage}[t]{0.465\linewidth}
    \vspace{0pt}\centering
    \subfloat[Grounding-to-Plan Transfer Gap. Measured on PAI-AV$^{*}$ safety-critical samples (Appendix~\ref{app:transfer_gap}).]{%
      \makebox[\linewidth][c]{\includegraphics[height=\figonepanelheight]{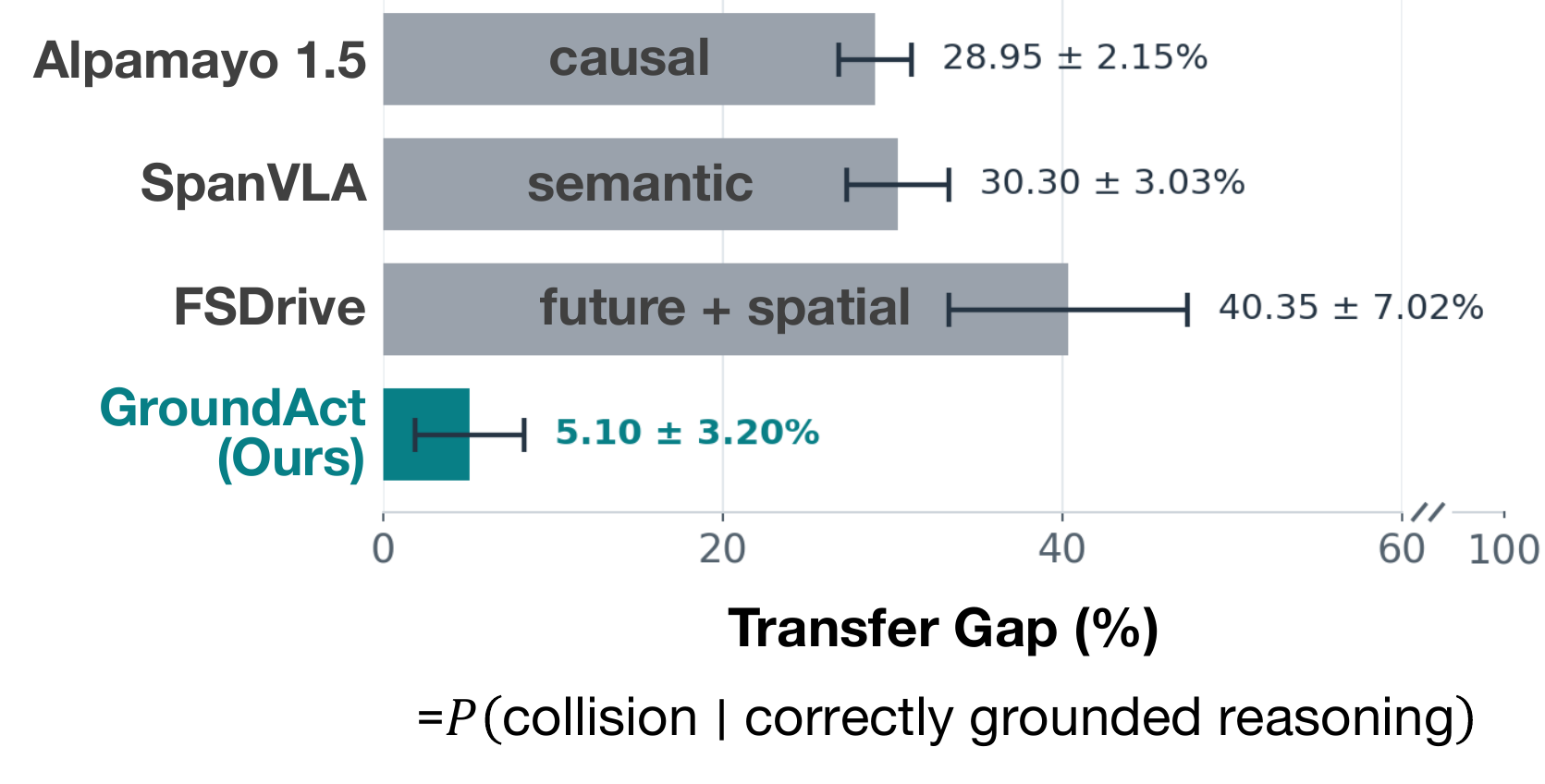}}%
      \label{fig1:a}%
    }
  \end{minipage}\hfill
  \begin{minipage}[t]{0.5\linewidth}
    \vspace{0pt}\centering
    \subfloat[Same Reasoning Trace, Different Actions Required. Blue: ego; orange/green: surrounding vehicles.]{%
      \makebox[\linewidth][c]{\includegraphics[height=\figonepanelheight]{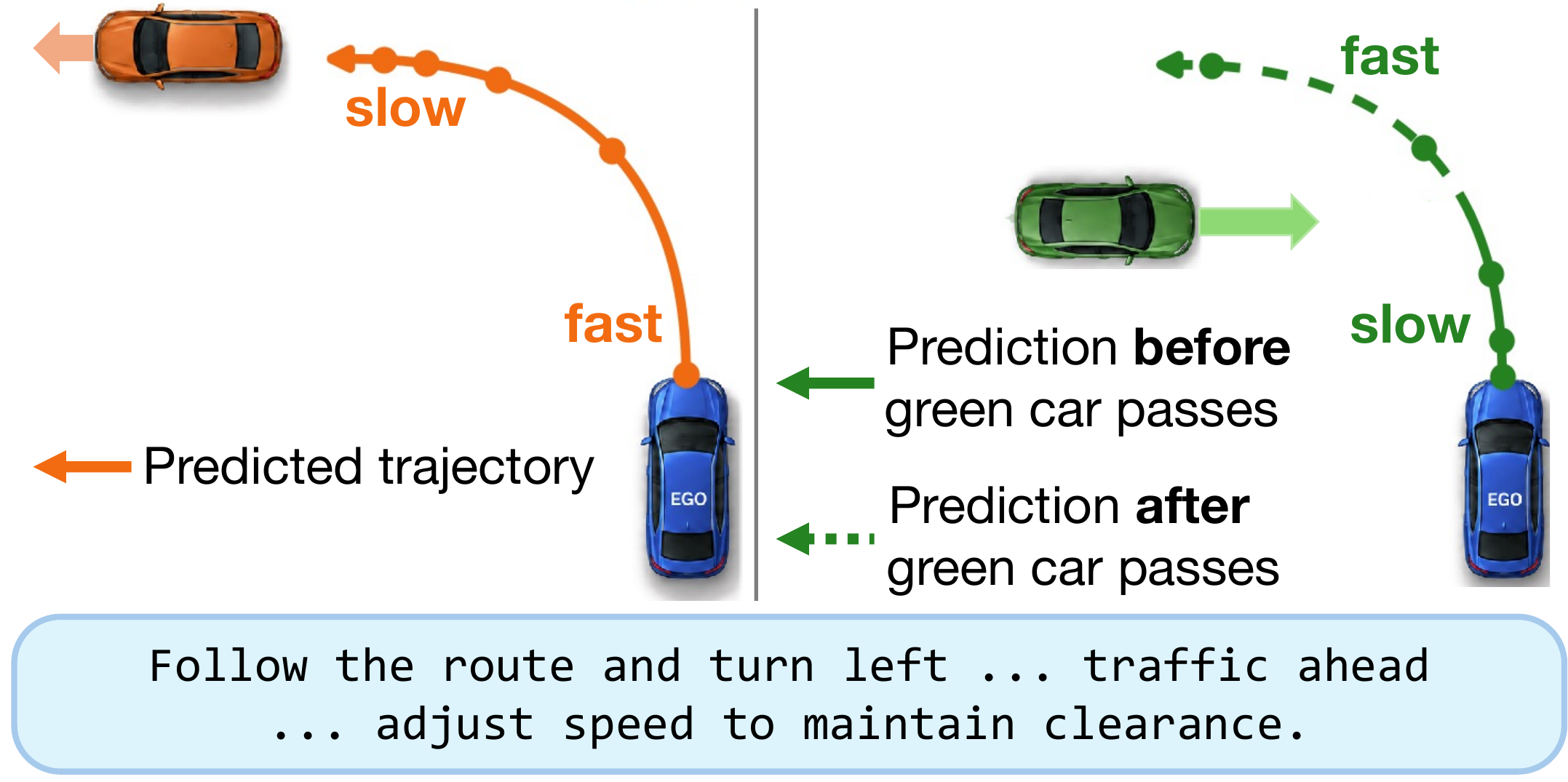}}%
      \label{fig1:b}%
    }
  \end{minipage}
  \caption{Grounded reasoning does not by itself ensure desirable driving outcomes.}
  \vspace{-8pt}
  \label{fig1}
  \endgroup
\end{figure}

Yet reasoning is not the endpoint of driving: the output is a plan.
Correctly grounded reasoning does not by itself ensure desirable driving outcomes, such as safety and comfort.
Our motivating audit measures this \emph{transfer gap}: the collision rate in the audited interactions, conditioned on correctly grounded reasoning (Figure~\ref{fig1}(a)).
This gap appears across causal, semantic, and spatiotemporal grounding.
Figure~\ref{fig1}(b) further illustrates how the same reasoning trace can arise in different driving contexts, even when the appropriate speed profiles differ.
Swapping the speed profiles between the two contexts can lead to safety-critical outcomes, such as collisions, despite correctly grounded reasoning in both cases.







This leads to our central question:
\begin{center}
    \emph{\textbf{What should groundedness mean when the output is an action?}}
\end{center}

\vspace{12pt}
Our answer begins from a physical-AI view of driving: the ego vehicle acts in a
world of physical entities and their interactions. Realizing a maneuver requires
accounting for where relevant entities are, how they move, and how the ego plan
interacts with them. We therefore choose the \emph{entity} as our grounding unit:
it provides a common physical referent linking what reasoning identifies as
relevant to what planning must account for. Our principle is to carry grounded
reasoning into action generation by preserving this physical connection, so that
the selected entities' interactions shape the evolving plan and improve the resulting driving behavior. We call this~\emph{grounded planning}.

In this work, we introduce \textbf{GroundAct}, whose \emph{Token-as-Reference} interface
interleaves generated reasoning content with references to continuous entity
states (Figure~\ref{fig:prior-vs-ours}). Resolving these references gives subsequent
reasoning and planning access to the same selected states. The action expert
retains the full scene context, while an explicit refinement pathway evaluates
the referenced states against the evolving ego proposal and uses their predicted
interactions to guide selective correction.


We evaluate GroundAct on three PAI-AV-based open-loop benchmarks (Normal, OOD, and Safety-Critical) and the AlpaSim closed-loop driving benchmark~\citep{physicalai_av, alpasim}. With 3.9B parameters, GroundAct demonstrates strong planning and safety performance across these benchmarks, remaining competitive with the 34B state-of-the-art driving foundation model~\citep{alpamayo_2s} and outperforming it on several metrics. Controlled ablations test whether reasoning-selected physical states remain consequential for planning by removing or replacing their references and by comparing reasoning-based selection against heuristic alternatives. Matched correction analyses further examine whether GroundAct corrects unsafe plans while preserving those that are already safe.

Our contributions are: (i)~\textbf{Grounded planning}, which establishes an
explicit and traceable path from the entity grounded in reasoning to its
consequence for action; (ii)~\textbf{Token-as-Reference}, which keeps
continuous entity state in the model while reasoning selects planning-critical
entities through lightweight symbolic references; and
(iii)~\textbf{Interaction-driven selective correction}, which uses the referenced entities' predicted interactions with the
evolving ego plan to guide selective refinement.

\FloatBarrier

%% file: sec/2_related_work.tex

\DeclareRobustCommand{\entitytoken}[1]{%
  \tikz[baseline=(tok.base)]{%
    \node[
      fill=entitygreen,
      text=white,
      rounded corners=6pt,
      inner xsep=4pt,
      inner ysep=1pt,
      font=\ttfamily\footnotesize
    ] (tok) {<|entity\_#1|>};%
  }%
}

\begin{figure} [tp]
    \centering
    \includegraphics[width=1\linewidth]{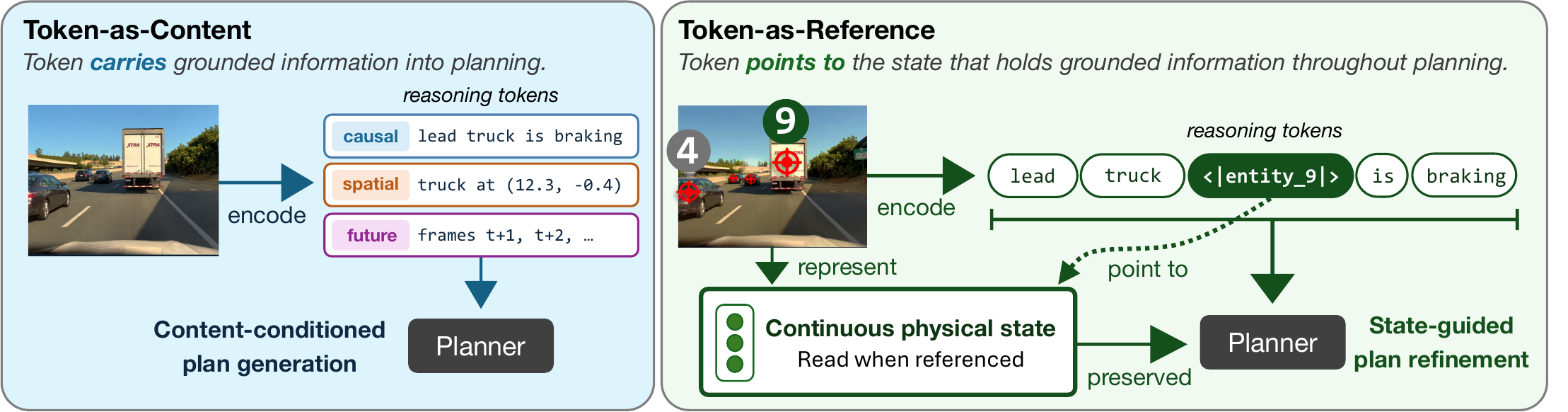}
    \vspace{-4pt}
    \caption{\textbf{Token-as-Content \textit{vs.} Token-as-Reference.}
        \textbf{[Left]} Grounded information is encoded in generated content.
        \textbf{[Right]} GroundAct retains this content path and adds entity-reference tokens \entitytoken{i}, each indexing a continuous entity state. These references keep the planning-critical entities selected during reasoning addressable throughout reasoning and planning. Only the referenced states enter the explicit refinement pathway.
        }
    \label{fig:prior-vs-ours}
\end{figure}

\vspace{-2pt}    
\section{Background and Related Work}
\label{sec:related-work}
\vspace{-6pt}
\paragraph{Grounded Reasoning as Token-as-Content.}
Driving VLAs ground their reasoning in three main forms, as illustrated on the left of Figure~\ref{fig:prior-vs-ours}.
\emph{Causally grounded} reasoning states the factors behind a decision in language,
as in Alpamayo-R1's causal chains~\citep{alpamayo} and in adaptive or self-reflective
reasoning before action~\citep{autovla,cf-vla,spanvla}.
\emph{Spatially grounded} reasoning carries positions: object coordinates written into the
reasoning text~\citep{token} or object boxes drawn into generated frames~\citep{futuresightdrive}.
\emph{Future-state grounded} reasoning incorporates predicted futures, represented as image tokens \citep{futuresightdrive}, latent tokens \citep{lcdrive,last-vla,driveworld-vla}, or compact dynamics tokens \citep{dynvla}.
Despite these differences, the three forms share one interface, which we call
\emph{Token-as-Content}: whatever is grounded is expressed in the generated tokens and
is consumed as such by subsequent reasoning or action generation. 
Grounding is thereby a property of the reasoning process; how that grounding should remain consequential for action generation is left unspecified.
Token-as-Reference draws inspiration from recent visual-grounding work:
\cite{v1} points to image patches and copies their embeddings into the reasoning stream, while \cite{tvg} serializes explicit points or boxes. 
Whereas these methods focus on grounding intermediate reasoning, GroundAct preserves the reference link from reasoning through downstream planning.
\vspace{-6pt}
\paragraph{Related Directions.}
GroundAct draws on two further lines of work, discussed in
Appendix~\ref{app:extended-related-work}. Prediction- and interaction-aware planners
show that structured prediction and physical interaction can shape planning directly
\citep{uniad,gameformer,p3,drivevla-w0,driveworld-vla}; GroundAct asks which entities
should receive that computation and how the choice stays connected to reasoning.
Guided and residual generative planners modify a pretrained generator without replacing
it \citep{flowmatching,scorediffusion,controlnet,actionintent}; GroundAct uses individual
referenced entities, rather than a global intent, as the mediator. 
Both lines make predictions or guidance signals consequential for planning; GroundAct
makes the \emph{reasoning's} grounding consequential, by keeping the selected entities
addressable until they act on the plan.

%% file: sec/3_method.tex
\section{GroundAct}
\label{sec:method}
\subsection{Overview}
\label{sec:method-overview}
\begin{figure}[tp]
    \centering
    \includegraphics[width=1\linewidth]{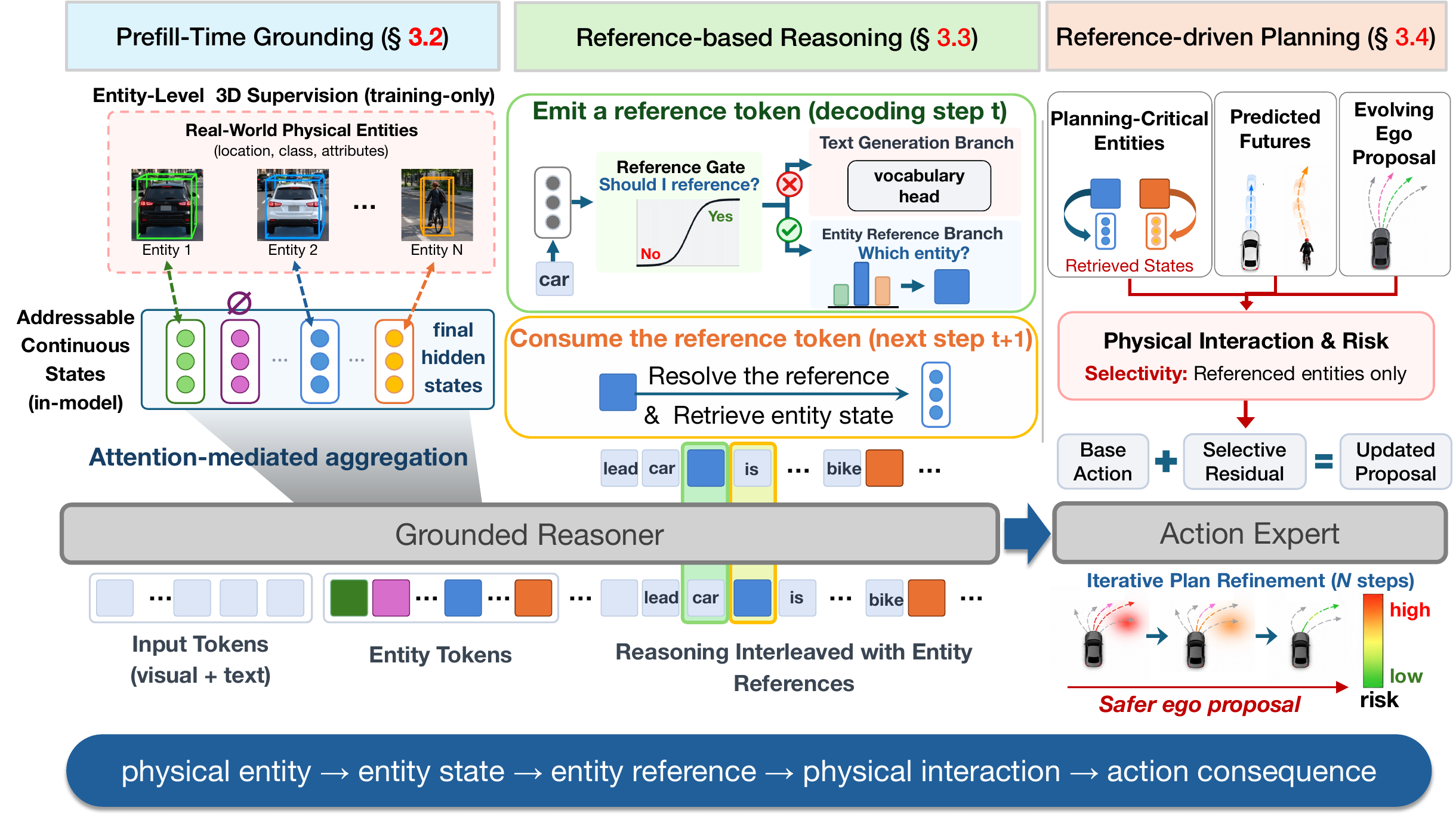}
    \caption{\textbf{GroundAct Architecture.}
        The three columns correspond to \S\ref{sec:ground} through \S\ref{sec:act}.
        The row below follows the computation from input tokens through the Physical AI
        reasoner to the action expert's refined plan, while the bottom bar summarizes
        the design invariant preserved throughout.}
    \label{fig:method-overview}
\end{figure}

GroundAct instantiates our three design choices: entity as the grounding unit, \emph{Token-as-Reference} as the grounding mechanism, and
interaction-driven correction for planning. These choices span continuous
multimodal encoding, discrete autoregressive reasoning, and iterative continuous
action generation (Figure~\ref{fig:method-overview}).

Figure~\ref{fig:method-overview} summarizes the computation. A physical entity is first
represented by a continuous, addressable state (\S\ref{sec:ground}).
During reasoning, an entity mention is associated with its corresponding state
in the current entity pool and expressed through a symbolic reference, without
serializing the underlying geometry into text (\S\ref{sec:reference}).
For planning, the referenced states are retrieved and explicitly evaluated
against the evolving ego proposal; their predicted physical
interactions then produce selective corrections to the generated action
(\S\ref{sec:act}). The representation changes across these computations, from
continuous state to discrete reference to interaction feature, while the
physical referent remains fixed. Section~\ref{sec:training-objectives}
describes how this correspondence is learned.

\subsection{Prefill-Time Grounding}
\label{sec:ground}
For reasoning and planning to refer to the same physical entity, the model first needs a
continuous state that is aligned with that entity and can be addressed later. Given
temporal observations and driving context $O$, we append $K$ entity tokens
$q_1,\ldots,q_K$ ($K{=}64$) to the multimodal prefill. Each acts as a learnable query that
attends to and aggregates temporal visual features together with the surrounding
context; its final hidden state

\[
    M_i = H_\theta(O,q_1,\ldots,q_K)[q_i],
\]

the Transformer output $H_\theta$ read at the position of $q_i$, is the entity state at
position $i$. The entity states $\mathcal{M}=\{M_i\}_{i=1}^{K}$ stay inside the model,
each addressable by its index $i$.

Each index $i$ is scene-local: the same position can represent different entities in
different scenes. Subsequent reasoning and planning must therefore bind to the state
an index currently addresses.

\subsection{Reference-based Reasoning}
\label{sec:reference}
Reasoning interprets the driving situation and the intended maneuver, and in doing so
singles out the planning-critical entities. To carry this choice into subsequent
reasoning and planning, \emph{Token-as-Reference} binds a discrete reference
$e_i=\langle|\mathrm{entity}\ i|\rangle$ to the continuous prefill state $M_i$.
The token names the entity; its physical information remains in the addressable state.

\paragraph{Reference Generation.}
At decoding step $t$, let $h_t$ be the reasoning hidden state and $y_{t+1}$ the next
token. A gate assigns probability mass $\pi_t$ to references, and an entity pointer
distributes it according to the correspondence between $h_t$ and the current states:

\begin{equation}
\label{eq:reference-generation}
    p(y_{t+1}=e_i\mid h_t,\mathcal{M})
    = \pi_t\,\operatorname{softmax}_i
      \!\left[\tau\cos(W_qh_t,W_kM_i)\right],
    \qquad \pi_t=\sigma(g(h_t)).
\end{equation}

Here $W_q,W_k$ are learned projections, $\tau>0$ controls score sharpness, and $g$ is
a scalar gate. The pointer softmax excludes detector-null slots whenever non-null
candidates are available. The vocabulary head distributes the remaining mass $1-\pi_t$ over
non-reference tokens. Here, the division of probability mass between vocabulary and reference outputs 
is motivated by the gated pointer-generator models~\citep{pointer-generator}.
This separates \emph{when} to reference from \emph{which}
entity to name within one next-token distribution.

\paragraph{Reference Resolution and Consumption.}
When $y_{t+1}=e_i$, resolving the address directly reads the original prefill state
$M_i$ from the pool. The next decoding step consumes this state through the input
embedding $u_{t+1}$:

\begin{equation}
\label{eq:reference-resolution}
\begin{aligned}
    \operatorname{Resolve}(e_i,\mathcal{M}) &= M_i,
    \qquad
    u_{t+1} &= (1-\alpha)b_{\mathrm{ref}}+\alpha\bigl(b_i+P(M_i)\bigr).
\end{aligned}
\end{equation}

Here $b_i$ combines the entity-query base embedding and slot identity,
$P$ projects and scales the retrieved state into the decoder input space, and
$\alpha\in[0,1]$ interpolates between this input and a shared reference embedding
$b_{\mathrm{ref}}$.
The decoder processes $u_{t+1}$ with its existing context to produce a new reasoning
hidden state; the retrieved $M_i$ remains the prefill representation. Planning reads
the same $M_i$ (\S\ref{sec:act}), preserving the selected physical referent across both
computations.

\subsection{Reference-driven Planning}
\label{sec:act}

A referenced entity does not prescribe an action: what it implies depends on where the
ego proposal (i.e., the planner's current candidate trajectory) is heading. Let
$\mathcal{R}(r)\subseteq\{1,\ldots,K\}$ denote the entities referenced in reasoning
sequence $r$. Only their states $\mathcal{M}_{\mathcal{R}}=\{M_i : i\in\mathcal{R}(r)\}$
enter explicit physical prediction and interaction modeling for residual correction, while the action expert---a
flow-matching trajectory generator conditioned on the reasoning context---keeps its full view of the scene. Reasoning thus decides where additional
physical computation is spent.

\paragraph{Interaction with the Evolving Proposal.}
In the continuous action mode, the action expert produces a trajectory by integrating a
learned vector field over a fixed number of solver steps. At step $k$, it evaluates the
base velocity $v_{\mathrm{base}}^{(k)}$ at the current flow state $x^{(k)}$; from the two
we extrapolate a sequence of normalized acceleration and curvature and decode it through
vehicle dynamics into the ego proposal
$\hat P^{(k)}$. In the forecast-based formulation, each referenced entity has a predicted motion
$\hat S_i^{(k)}$ over the planning horizon (in the joint variant, conditioned on the
current proposal), and the proposal-relative interaction

\[
    C_i^{(k)} = \operatorname{Interact}\!\left(\hat P^{(k)},\hat S_i^{(k)}\right),
    \qquad i\in\mathcal{R}(r),
\]

encodes quantities such as clearance, relative motion, and validity over that horizon.
Because the proposal changes at every step, so do these interactions; $k$ indexes
refinement of the whole proposal, not time within the horizon.

\paragraph{Selective Correction.}
In this general formulation, a permutation-invariant set encoder aggregates the referenced states, their forecasts,
and their interactions, and a residual adapter $R_\phi$ maps the result, together with
the current state and base velocity, to a correction of the base field:

\[
    z_{\mathcal{R}}^{(k)}
    = \operatorname{SetEnc}\!\left(\left\{M_i,\hat S_i^{(k)},C_i^{(k)}\right\}_{i\in\mathcal{R}(r)}\right),
    \qquad
    v_{\mathrm{GroundAct}}^{(k)}
    = v_{\mathrm{base}}^{(k)}
    + \gamma^{(k)}\,R_\phi\!\left(x^{(k)},v_{\mathrm{base}}^{(k)},z_{\mathcal{R}}^{(k)}\right),
\]

where $\gamma^{(k)}$ combines a risk latch, activated by predicted overlap for the remaining solver steps, with a flow-time schedule. The solver integrates the corrected field, and the new proposal is evaluated again. The correction is defined jointly over the referenced set, and the pathway is bypassed (i.e., recovering the base field) when no valid entity is referenced or the gate is inactive.
Our implementation predicts clearance from four observed frames, the ego proposal, and the states of valid referenced entities; thus, clearance conditions the residual.
\input{sec/3_training_objectives}

%% file: sec/3_training_objectives.tex
\begingroup
\subsection{Training Objectives And Optimization}
\label{sec:training-objectives}

Training preserves physical correspondence from entity states to references and
selective action correction. The principal objectives below apply to different parameter
sets and optimization intervals.

\paragraph{Entity States.}
Entity-level 3D supervision establishes physical correspondence for the entity
states. Prediction heads decode an object category, including a null class, and
an oriented BEV box from each $M_i$. Hungarian matching assigns predictions one-to-one to annotated
objects; unmatched queries receive the null target. Class/null prediction and
box regression are combined with language modeling:

\begin{equation}
\mathcal L_{\mathrm{entity}}
=\lambda_{\mathrm{cls}}\mathcal L_{\mathrm{cls}}
+\lambda_{\mathrm{box}}\mathcal L_{\mathrm{box}}
+\lambda_{\mathrm{LM}}\mathcal L_{\mathrm{LM}}.
\label{eq:train-entity}
\end{equation}

Queries and prediction heads first warm up with frozen backbones. Joint
adaptation then unfreezes the backbones and includes language supervision.

\paragraph{Physical References.}
Offline annotations link accepted referring expressions in reasoning to
physical objects. At each training pass, those objects are rematched to
class-consistent states in the current entity pool by BEV-center distance,
so reference targets follow physical correspondences rather than fixed indices.
After pointer alignment, decoder adaptation combines masked token prediction
$\mathcal L_{\mathrm{text}}$, matched-entity association
$\mathcal L_{\mathrm{assoc}}$, and reference timing $\mathcal L_{\mathrm{gate}}$:

\begin{equation}
\mathcal L_{\mathrm{reason}}
=\lambda_{\mathrm{text}}\mathcal L_{\mathrm{text}}
+\lambda_{\mathrm{assoc}}\mathcal L_{\mathrm{assoc}}
+\lambda_{\mathrm{gate}}\mathcal L_{\mathrm{gate}}.
\label{eq:train-reasoning}
\end{equation}


\paragraph{Selective Correction.}
The base action expert is trained with conditional flow matching and then
frozen for physical prediction and residual learning.
For forecast-based prediction, annotated entity motion supervises the forecasts through
$\mathcal L_{\mathrm{future}}$ (Appendix~\ref{app:training-recipe},
\Eqref{eq:train-future}). As an alternative, physical supervision targets proposal-relative clearance 
computed from recorded entity motion, with higher weight near contact. 
In addition, offline planner screens for target-clearance gain, retained progress,
and no newly overlapping entities define residual targets relative to the frozen planner's field at the same stored
flow state. Distillation fits the applied correction with
$\mathcal L_{\mathrm{field}}$ and the ungated residual with
$\mathcal L_{\mathrm{raw}}$, penalizes raw corrections in risk-inactive states
with $\mathcal L_{\mathrm{off}}$, and regularizes their consistency and magnitude with $\mathcal L_{\mathrm{reg}}$:

\begin{equation}
\mathcal L_{\mathrm{res}}
=\lambda_{\mathrm{field}}\mathcal L_{\mathrm{field}}
+\lambda_{\mathrm{raw}}\mathcal L_{\mathrm{raw}}
+\lambda_{\mathrm{off}}\mathcal L_{\mathrm{off}}
+\mathcal L_{\mathrm{reg}}.
\label{eq:train-preservation}
\end{equation}

The residual fitting uses ground-truth-derived interaction features;
inference uses predicted physical information. Appendix~\ref{app:training-recipe}
defines the component losses, teacher targets, and optimization scope.
\endgroup

%% file: sec/4_evaluation.tex
\section{Evaluation}
\label{sec:eval}

\subsection{Experimental Setup}
\label{sec:eval-setup}

\definecolor{evalrankfirstbg}{HTML}{BFE3DC}
\definecolor{evalranksecondbg}{HTML}{F7D9BE}
\definecolor{evalrankthirdbg}{HTML}{DDD5EF}
\newcommand{\evalrankfirst}[1]{\cellcolor{evalrankfirstbg}#1}
\newcommand{\evalranksecond}[1]{\cellcolor{evalranksecondbg}#1}
\newcommand{\evalrankthird}[1]{\cellcolor{evalrankthirdbg}#1}

\begin{table}[t]
  \centering
  \caption{\textbf{Open-Loop Driving Performance by Grounding Paradigm.}
      Generation mode: \evalmodeD{} discrete; \evalmodeC{} continuous;
      \colorbox{evalrankfirstbg}{best}, \colorbox{evalranksecondbg}{second}, and \colorbox{evalrankthirdbg}{third} distinct values per column.}
  \label{tab:eval-open-loop}
  \vspace{2pt}
  \begingroup
  \fontsize{7}{8.4}\selectfont
  \setlength{\tabcolsep}{0.6pt}
  \renewcommand{\arraystretch}{1.2}

  \begin{tabular*}{\linewidth}{@{}p{2.5pt}@{\hskip 2pt} p{55pt} p{76pt} >{\centering\arraybackslash}p{18pt} >{\centering\arraybackslash}p{20pt} @{\extracolsep{\fill}} cccccc cccccc @{}}
    \toprule
    & & & & & \multicolumn{6}{c}{\cellcolor{dsnormal}\textbf{PAI-AV Normal} ($n{=}4K$)} & \multicolumn{6}{c}{\cellcolor{dsood}\textbf{PAI-AV OOD} ($n{=}290$)} \\
    \cmidrule(lr){6-11}\cmidrule(l){12-17}
    & Model & Token represents & Size & Mode
      & \multicolumn{3}{c}{ADE} & \multicolumn{2}{c}{Collision} & \shortstack{Corner\\Dist.}
      & \multicolumn{3}{c}{ADE} & \multicolumn{2}{c}{Collision} & \shortstack{Corner\\Dist.} \\
    & & & & & \hz{2.5\,s} & \hz{5.0\,s} & \hz{6.4\,s} & \hz{2.5\,s} & \hz{5.0\,s} & \hz{6.4\,s}
            & \hz{2.5\,s} & \hz{5.0\,s} & \hz{6.4\,s} & \hz{2.5\,s} & \hz{5.0\,s} & \hz{6.4\,s} \\
    \midrule
    \addlinespace[3pt]
    \rowcolor{parcontenttint}
    \cellcolor{parcontent} & \multicolumn{16}{@{}l}{\textbf{\textcolor{parcontent}{Token-as-Content}}}\\
    \cellcolor{parcontent} & \mrow FSDrive$^{\dagger}$ & \gin{spatio-temporal states} & 2.1B & \evalmodeD{} & 0.417 & 1.809 & 3.112 & 2.943 & 10.721 & 3.231 & 0.546 & 2.145 & 3.404 & 3.549 & 20.469 & 3.480 \\
    \cellcolor{parcontent} & \mrow Alpamayo-R1 & \gin{causes} & 10B & \evalmodeD{} & 0.551 & 2.612 & 4.417 & 4.269 & 13.443 & 4.599 & 0.596 & 2.691 & 4.402 & 5.805 & 24.844 & 4.568 \\
    \cellcolor{parcontent} & \mrow Alpamayo-R1 & \gin{causes} & 10B & \evalmodeC{} & 0.535 & 2.263 & 3.802 & 3.714 & 12.472 & 3.945 & 0.541 & 2.295 & 3.751 & 4.182 & 21.036 & 3.872 \\
    \cellcolor{parcontent} & \mrow AutoVLA$^{\dagger}$ & \gin{semantic knowledge} & 2.1B & \evalmodeD{} & 0.555 & 1.780 & 2.848 & 2.557 & 9.014 & 2.910 & 0.563 & 1.973 & 3.096 & 2.809 & 14.170 & 3.148 \\
    \cellcolor{parcontent} & \mrow LCDrive & \gin{action-conditioned futures} & 0.6B & \evalmodeD{} & 0.332 & 1.503 & 2.615 & \evalrankthird{2.292} & 9.100 & 2.676 & 0.475 & 1.994 & 3.268 & 2.809 & 17.353 & 3.330 \\
    \cellcolor{parcontent} & \mrow Alpamayo 1.5 & \gin{causes} & 10B & \evalmodeD{} & 0.425 & 1.881 & 3.163 & 3.311 & 10.951 & 3.279 & 0.522 & 2.148 & 3.412 & 3.995 & 21.223 & 3.513 \\
    \cellcolor{parcontent} & \mrow Alpamayo 1.5 & \gin{causes} & 10B & \evalmodeC{} & 0.575 & 2.247 & 3.647 & 3.952 & 12.806 & 3.770 & 0.585 & 2.240 & 3.528 & 4.370 & 21.848 & 3.627 \\
    \cellcolor{parcontent} & \mrow SpanVLA$^{\dagger}$ & \gin{semantic knowledge} & 3.8B & \evalmodeC{} & 0.496 & 1.818 & 2.844 & 2.531 & 8.339 & 2.931 & 0.482 & 2.059 & 3.167 & \evalrankthird{2.310} & 14.232 & 3.532 \\
    \cellcolor{parcontent} & \mrow Alpamayo 2 Super & \gin{causes} & 34B & \evalmodeD{} & 0.345 & 1.338 & 2.245 & 2.591 & \evalrankthird{6.028} & 2.304 & 0.470 & 1.696 & 2.683 & 2.871 & \evalranksecond{9.613} & 2.756 \\
    \cellcolor{parcontent} & \mrow Alpamayo 2 Super & \gin{causes} & 34B & \evalmodeC{} & \evalranksecond{0.299} & \evalranksecond{1.304} & \evalrankthird{2.214} & \evalrankfirst{2.097} & \evalrankfirst{5.473} & \evalranksecond{2.260} & \evalranksecond{0.391} & \evalranksecond{1.606} & \evalranksecond{2.601} & \evalrankfirst{2.060} & \evalrankfirst{8.864} & \evalranksecond{2.645} \\
    \addlinespace[3pt]
    \rowcolor{parreferencetint}
    \cellcolor{parreference} & \multicolumn{16}{@{}l}{\textbf{\textcolor{parreference}{Token-as-Reference (Ours)}}} \\
    \cellcolor{parreference} & \mrow \ours & \gin{entity-state addresses} & 3.9B & \evalmodeD{} & \evalrankfirst{0.266} & \evalrankfirst{1.204} & \evalrankfirst{2.091} & \evalranksecond{2.149} & \evalranksecond{5.967} & \evalrankfirst{2.138} & \evalrankfirst{0.371} & \evalrankfirst{1.545} & \evalrankfirst{2.540} & \evalranksecond{2.185} & 11.923 & \evalrankfirst{2.592} \\
    \cellcolor{parreference} & \mrow \ours & \gin{entity-state addresses} & 3.9B & \evalmodeC{} & \evalrankthird{0.325} & \evalrankthird{1.332} & \evalranksecond{2.210} & 2.474 & 6.873 & \evalrankthird{2.278} & \evalrankthird{0.459} & \evalrankthird{1.692} & \evalrankthird{2.680} & \evalrankthird{2.310} & \evalrankthird{10.861} & \evalrankthird{2.747} \\
    \addlinespace[1pt]
    \bottomrule
  \end{tabular*}

  \vspace{5pt}
  \begin{tabular*}{\linewidth}{@{}p{2.5pt}@{\hskip 2pt} p{55pt} p{76pt} >{\centering\arraybackslash}p{18pt} >{\centering\arraybackslash}p{20pt} @{\extracolsep{\fill}} cccccccc @{}}
    \toprule
    & & & & & \multicolumn{8}{c}{\cellcolor{dssafety}\textbf{PAI-AV$^{*}$ Safety-Critical} ($n{=}512$)} \\
    \cmidrule(l){6-13}
    & Model & Token represents & Size & Mode
      & \multicolumn{3}{c}{ADE} & \multicolumn{2}{c}{Collision} & \multicolumn{2}{c}{Off-road} & \shortstack{Corner\\Dist.} \\
    & & & & & \hz{2.5\,s} & \hz{5.0\,s} & \hz{6.4\,s} & \hz{2.5\,s} & \hz{5.0\,s} & \hz{2.5\,s} & \hz{5.0\,s} & \hz{6.4\,s} \\
    \midrule
    \addlinespace[3pt]
    \rowcolor{parcontenttint}
    \cellcolor{parcontent} & \multicolumn{12}{@{}l}{\textbf{\textcolor{parcontent}{Token-as-Content}}} \\
    \cellcolor{parcontent} & \mrow FSDrive$^{\dagger}$ & \gin{spatio-temporal states} & 2.1B & \evalmodeD{} & 0.533 & 2.249 & 3.823 & 3.797 & 20.161 & 5.712 & 28.394 & 3.992 \\
    \cellcolor{parcontent} & \mrow Alpamayo-R1 & \gin{causes} & 10B & \evalmodeD{} & 0.977 & 4.084 & 6.584 & 6.770 & 29.490 & 12.695 & 37.565 & 6.939 \\
    \cellcolor{parcontent} & \mrow Alpamayo-R1 & \gin{causes} & 10B & \evalmodeC{} & 1.014 & 4.054 & 6.521 & 7.230 & 28.840 & 14.388 & 38.411 & 6.872 \\
    \cellcolor{parcontent} & \mrow AutoVLA$^{\dagger}$ & \gin{semantic knowledge} & 2.1B & \evalmodeD{} & 0.570 & 1.907 & 3.027 & 1.465 & 14.290 & 4.297 & 17.383 & 3.069 \\
    \cellcolor{parcontent} & \mrow LCDrive & \gin{action-conditioned futures} & 0.6B & \evalmodeD{} & \evalranksecond{0.384} & 1.585 & 2.633 & 1.074 & 12.370 & 1.530 & 9.668 & 2.709 \\
    \cellcolor{parcontent} & \mrow Alpamayo 1.5 & \gin{causes} & 10B & \evalmodeD{} & 0.782 & 3.278 & 5.281 & 6.413 & 29.330 & 10.286 & 36.198 & 5.548 \\
    \cellcolor{parcontent} & \mrow Alpamayo 1.5 & \gin{causes} & 10B & \evalmodeC{} & 0.865 & 3.384 & 5.396 & 6.152 & 29.460 & 10.221 & 37.402 & 5.665 \\
    \cellcolor{parcontent} & \mrow SpanVLA$^{\dagger}$ & \gin{semantic knowledge} & 3.8B & \evalmodeC{} & 0.561 & 2.213 & 3.182 & 1.302 & 15.625 & 2.539 & 17.643 & 5.195 \\
    \cellcolor{parcontent} & \mrow Alpamayo 2 Super & \gin{causes} & 34B & \evalmodeD{} & 0.528 & 1.768 & 2.797 & 1.335 & \evalrankthird{7.585} & 3.092 & \evalrankthird{7.878} & 2.887 \\
    \cellcolor{parcontent} & \mrow Alpamayo 2 Super & \gin{causes} & 34B & \evalmodeC{} & \evalranksecond{0.384} & \evalrankfirst{1.515} & \evalrankfirst{2.491} & \evalrankfirst{0.390} & \evalranksecond{7.420} & \evalrankthird{1.497} & \evalranksecond{7.585} & \evalrankfirst{2.558} \\
    \addlinespace[3pt]
    \rowcolor{parreferencetint}
    \cellcolor{parreference} & \multicolumn{12}{@{}l}{\textbf{\textcolor{parreference}{Token-as-Reference (Ours)}}} \\
    \cellcolor{parreference} & \mrow \ours & \gin{entity-state addresses} & 3.9B & \evalmodeD{} & \evalrankfirst{0.382} & \evalranksecond{1.528} & \evalrankthird{2.506} & \evalrankthird{0.553} & \evalrankfirst{7.096} & \evalranksecond{1.107} & \evalrankfirst{6.868} & \evalrankthird{2.580} \\
    \cellcolor{parreference} & \mrow \ours & \gin{entity-state addresses} & 3.9B & \evalmodeC{} & \evalrankthird{0.431} & \evalrankthird{1.565} & \evalranksecond{2.501} & \evalranksecond{0.488} & 9.668 & \evalrankfirst{1.042} & 8.496 & \evalranksecond{2.571} \\
    \bottomrule
  \end{tabular*}
  \endgroup
\end{table}

\paragraph{Models and Comparisons.}
We instantiate \method{} with a Qwen3-VL-2B vision--language backbone~\citep{qwen3-vl-2b}
and a flow-matching action expert (2.1B and 1.8B parameters; Table~\ref{tab:eval-open-loop}). 
Actions are generated either as discrete action tokens from the VLM backbone or as a continuous trajectory from the action expert; the selective correction of \S\ref{sec:act} modifies the expert's vector field, so it is active only in the continuous mode.
We nonetheless report discrete generation in the open-loop evaluation, which isolates the effect of entity grounding on downstream reasoning from that of the correction.
We compare against Alpamayo-R1~\citep{alpamayo}, Alpamayo 1.5~\citep{alpamayo15},
Alpamayo 2 Super~\citep{alpamayo_2s}, FSDrive$^{\dagger}$~\citep{futuresightdrive}, LCDrive~\citep{lcdrive},
AutoVLA$^{\dagger}$~\citep{autovla}, and SpanVLA$^{\dagger}$~\citep{spanvla}. The Alpamayo models
and LCDrive are evaluated from their released checkpoints without fine-tuning.
FSDrive$^{\dagger}$, AutoVLA$^{\dagger}$ and SpanVLA$^{\dagger}$ are the matched controls: each method’s defining component is re-implemented on our backbone and fine-tuned, without subsequent reinforcement learning, on the
same 140.5K PAI-AV training samples as \method{}.
See Appendix~\ref{app:experimental-setup-details} for further details.

\paragraph{Driving Evaluation Protocols.}
Open-loop evaluation uses the PAI-AV Normal (4K samples) and OOD (290 samples) splits, along with 512 high-risk samples from PAI-AV$^{*}$ collected under the same data configuration (Appendix~\ref{app:safety-critical-distribution}).
All models receive four temporally ordered front-wide camera frames ($t{-}0.3$\,s, $t{-}0.2$\,s, $t{-}0.1$\,s, $t$) and generate six candidate trajectories over a 6.4\,s horizon.
Following LCDrive~\citep{lcdrive}, we report ADE, corner distance, collision rate, and off-road rate averaged over the six candidates.
Closed-loop evaluation uses 913 scenes from the AlpaSim `public\_2601' test suite~\citep{alpasim}, with the same four-frame input.
We report progress, three safety metrics, and two diagnostic distances (Table~\ref{tab:eval-closed-loop}; Appendix~\ref{app:closed-loop-metrics}).

\subsection{Driving Performance}
\label{sec:eval-driving}
\paragraph{Open-Loop Planning.}

Table~\ref{tab:eval-open-loop} evaluates \method{} under both discrete and continuous generation. Across all three datasets, \method{} achieves the best or comparable performance on nearly all metrics, despite being substantially smaller than Alpamayo 2 Super. The discrete variant operates solely with the 2.1B VLM, rather than the full 3.4B model. Here, entity references retrieve the corresponding physical states into the reasoning process, yielding strong trajectory accuracy across all three datasets and long-horizon safety on the safety-critical set.
The continuous variant additionally incorporates state-based plan correction through the action expert. Its performance remains competitive but slightly trails Alpamayo 2 Super on several metrics. Table~\ref{tab:eval-grounding-ablation}~attributes this gap to the flow-matching head (i.e., action expert trained from scratch on substantially less data and without Alpamayo's RL post-training) rather than to the correction itself. Together, these results show the benefit of entity grounding in both reasoning and continuous plan refinement.

\definecolor{evaldetect}{HTML}{EAF1F6}
\definecolor{evalrefer}{HTML}{F1EDF6}
\begin{table*}[t]
\centering
\begingroup
\caption{\textbf{Closed-Loop Driving Performance on AlpaSim.}
Each model uses its representative generation mode: discrete generation for FSDrive, AutoVLA, and LCDrive, and continuous generation otherwise; \colorbox{evalrankfirstbg}{best}, \colorbox{evalranksecondbg}{second}, and \colorbox{evalrankthirdbg}{third} distinct values for Progress and Safety Metrics. Metric definitions are provided in Appendix~\ref{app:closed-loop-metrics}.}
\vspace{2pt}
\label{tab:eval-closed-loop}

\fontsize{7.2}{8.4}\selectfont
\setlength{\tabcolsep}{2.0pt}
\renewcommand{\arraystretch}{1.2}

\begin{tabular*}{\textwidth}{@{}l @{\extracolsep{\fill}} c c c c c c c@{}}
\toprule
\multicolumn{3}{c}{} & \multicolumn{3}{>{\columncolor{evaldetect}}c}{\textbf{Safety Metrics} $\downarrow$}
& \multicolumn{2}{>{\columncolor{evalrefer}}c}{\textbf{Additional Analysis (m)}} \\
\cmidrule(lr){4-6}\cmidrule(l){7-8}
\textbf{Model}
& \textbf{\# Params}
& \textbf{Progress} $\uparrow$
& \shortstack{At-fault\\collision}
& Off-road
& \shortstack{Wrong\\lane}
& \shortstack{Distance to\\GT Traj.}
& \shortstack{Min. obstacle\\distance} \\
\midrule
FSDrive$^{\dagger}$~\citep{futuresightdrive} & 2.1B & 0.604 & 0.193 & 0.271 & \evalrankthird{0.210} & 2.577 & 4.447 \\
Alpamayo-R1~\citep{alpamayo} & 10B & 0.682 & 0.156 & 0.189 & 0.295 & 2.781 & 4.649 \\
AutoVLA$^{\dagger}$~\citep{autovla} & 2.1B & \evalrankthird{0.706} & \evalrankthird{0.118} & 0.185 & 0.211 & 2.544 & 3.989 \\
LCDrive~\citep{lcdrive} & 0.6B & 0.639 & 0.259 & 0.263 & \evalrankfirst{0.189} & 2.825 & 3.871 \\
Alpamayo 1.5~\citep{alpamayo15} & 10B & 0.682 & 0.166 & \evalrankthird{0.155} & 0.297 & 2.843 & 4.407 \\
SpanVLA$^{\dagger}$~\citep{spanvla} & 3.8B & 0.696 & 0.156 & 0.184 & \evalranksecond{0.207} & 2.896 & 3.794 \\
Alpamayo 2 Super~\citep{alpamayo_2s} & 34B & \evalranksecond{0.713} & \evalrankfirst{0.084} & \evalrankfirst{0.147} & 0.220 & 2.799 & 3.721 \\ \midrule
\textbf{\method{}} & 3.9B & \evalrankfirst{0.752} & \evalranksecond{0.105} & \evalranksecond{0.153} & \evalranksecond{0.207} & 2.511 & 3.641 \\
\bottomrule
\end{tabular*}
\endgroup
\end{table*}

\paragraph{Closed-Loop Driving.}
 In Table~\ref{tab:eval-closed-loop}, \method{}, with the correction of \S\ref{sec:act} active, attains the highest progress, ahead of the much larger, RL post-trained Alpamayo 2 Super, and is the only model in the top two on progress and all three safety metrics. It is better than or tied with every matched control, which shares its backbone and training data, on each of these metrics, including SpanVLA$^{\dagger}$ at comparable size. It is also comparable to Alpamayo 2 Super, with only marginal gaps in at-fault collision and off-road performance. The diagnostic columns indicate that this safety does not come from conservative margins: \method{} deviates least from the GT path despite driving furthest and has the smallest clearance to other actors, consistent with selective correction (Table~\ref{tab:reference-selectivity}(b)), whereas the two earlier Alpamayo models keep wider clearance yet show the highest wrong-lane rates.

\begin{figure}[t]
  \centering
  \includegraphics[width=\linewidth]{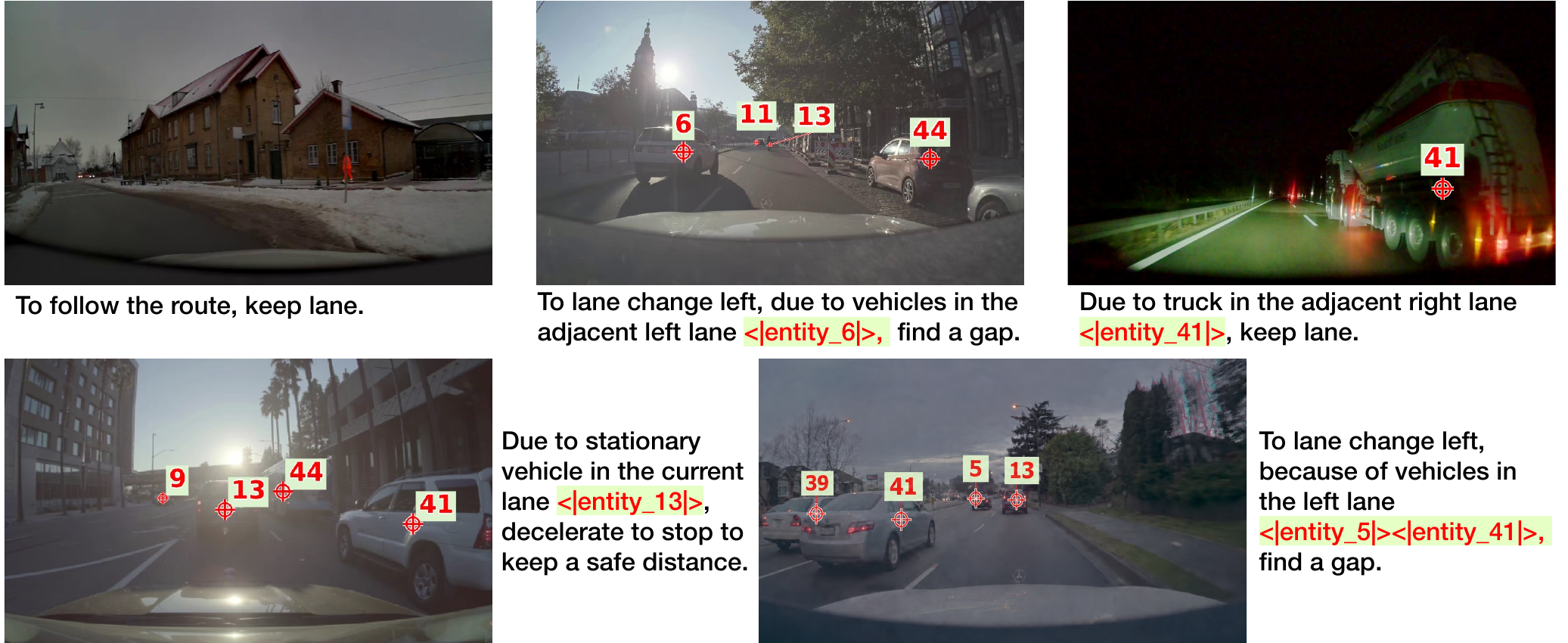}
    \caption{Reference-Based Reasoning with Varying Entity References.}
  \label{fig:gounded_reasoning}
\end{figure}

\begingroup
    \setlength{\textfloatsep}{10pt}
        \begin{figure}[pt]
        \centering
        \definecolor{evalabquality}{RGB}{235,244,249}
        \definecolor{evalabsafety}{RGB}{244,239,248}
        \newsavebox{\ablTableBox}
        \begin{lrbox}{\ablTableBox}%
        \fontsize{8.5}{8.5}\selectfont
        \renewcommand{\arraystretch}{0.95}%
        \begin{tabular}{@{}lcccc@{}}
        \toprule
        & \shortstack{VLM\strut\\backbone\strut} & \shortstack{Spatial\strut\\grounding\strut}
        & \shortstack{Grounded\strut\\reasoning\strut} & \shortstack{\textbf{GroundAct}\strut\\\textbf{(Ours)}\strut} \\
        \midrule
        Ground    & -- & $\checkmark$ & $\checkmark$ & $\checkmark$ \\
        Reference & -- & -- & $\checkmark$ & $\checkmark$ \\
        Act       & -- & -- & -- & $\checkmark$ \\
        \midrule
        \multicolumn{5}{@{}l@{}}{\cellcolor{evalabquality}\strut\textbf{Trajectory (m)}$\downarrow$} \\
        ADE 2.5\,s & 0.551 & 0.479 & \underline{0.463} & \textbf{0.431} \\
        ADE 5.0\,s & 2.308 & 1.893 & \underline{1.651} & \textbf{1.565} \\
        ADE 6.4\,s & 3.498 & 2.985 & \underline{2.581} & \textbf{2.501} \\
        \midrule
        \multicolumn{5}{@{}l@{}}{\cellcolor{evalabsafety}\strut\textbf{Safety (\%)}$\downarrow$} \\
        Coll.\ 2.5\,s   & 1.465  & 1.237  & \underline{0.716}  & \textbf{0.488} \\
        Coll.\ 5.0\,s   & 17.741 & 13.639 & \underline{11.979} & \textbf{9.668} \\
        Off-road 2.5\,s & 2.637  & 1.855  & \underline{1.302}  & \textbf{1.042} \\
        Off-road 5.0\,s & 19.857 & 17.415 & \underline{12.988} & \textbf{8.496} \\
        \bottomrule
        \end{tabular}%
        \end{lrbox}%
        \newlength{\ablTableHt}%
        \setlength{\ablTableHt}{\dimexpr\ht\ablTableBox+\dp\ablTableBox\relax}%
        \begin{minipage}[t]{0.46\linewidth}
        \vspace{0pt}\centering
        \begingroup
        \expandafter\def\csname @captype\endcsname{table}
        \caption{\textbf{Make Grounding Consequential.} Continuous generation; \textbf{Best}; \underline{second-best}.}
        \label{tab:eval-grounding-ablation}
        \vspace{6pt}
        \scalebox{0.85}{\usebox{\ablTableBox}}
        \endgroup
        \end{minipage}\hspace{0.03\linewidth}%
        \begin{minipage}[t]{0.47\linewidth}
        \vspace{8pt}\centering
        \includegraphics[height=\ablTableHt]{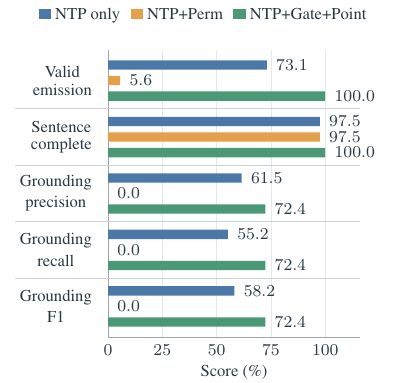}
        \vspace{-10pt}
        \caption{\hyphenpenalty=10000 \textbf{Learn to Reference.}}
        \label{fig:eval-reference-objectives}
        \end{minipage}
        
        \par
        \begin{minipage}{\linewidth}
        \end{minipage}
    \end{figure}
\endgroup

\subsection{Ablation and Analysis}
\label{sec:eval-reference-analysis}
\label{sec:eval-ablation}
\vspace{-3pt}
We examine three aspects of \method{}’s grounding pathway: whether the grounding connection should be preserved through action refinement, whether the model learns current entity–reference associations, and whether reasoning-selected references remain consequential for planning and support selective correction.

\vspace{-3pt}
\paragraph{Grounding Connection.}
Table~\ref{tab:eval-grounding-ablation} examines the value of carrying
grounded information through reasoning into action refinement.
With a shared backbone and continuous-generation protocol, grounded
reasoning improves on spatial grounding alone, and interaction-driven
correction further reduces trajectory error, collision rates, and
off-road rates. These results support the combined design, without
isolating each component's standalone contribution.

\vspace{-4pt}
\paragraph{Learning Current Entity Associations.}
Under the reported evaluation, successful language generation can coexist
with failed entity referencing
(Figure~\ref{fig:eval-reference-objectives}). The comparison supports
explicitly supervising both when to refer and which current entity state to
select, accommodating changing entity--token associations.

\vspace{-4pt}
\paragraph{Reference Dependence and Selection.}
The intervention audit shows that references are consequential for planning,
with evidence of safety utility on the Safety cohort
(Table~\ref{tab:reference-selectivity}(a)). It does not establish that reasoning-based selection is safer than the tested heuristics, or equivalent to them.

\begin{figure}[t]
  \centering
    \begin{minipage}{\linewidth}
      \expandafter\def\csname @captype\endcsname{table}
      \centering
      \caption{\textbf{Reference Dependence and Selective Correction.}
        (a) Collision-rate increase relative to Ours (pp).
        \emph{Matched distractor}: another same-class entity at a similar distance,
        with at least 1\,m clearance from the base plan.
        \emph{Highest predicted risk}: the entity with the highest predicted risk
        to the ego plan.
        (b) \emph{Preserved ($X$\,m)}: percentage of initially safe plans whose
        maximum time-aligned distance from the base plan is at most $X$\,m
        over 6.4\,s.}
      \vspace{8pt}
    
      \label{tab:reference-selectivity}
      \begingroup
      \fontsize{8.5}{8.5}\selectfont
      \setlength{\tabcolsep}{1.5pt}
      \renewcommand{\arraystretch}{1.10}
      \begin{minipage}{0.92\linewidth}
      \begin{minipage}[t]{0.52\linewidth}
        \begin{tabular*}{\linewidth}{@{\extracolsep{\fill}}lcc@{}}
          \toprule
          \rowcolor{dssafety}
          \multicolumn{3}{l}{\textbf{(a) Reference-to-Action Dependence}}\\
          Reference Selection & PAI-AV* Safety & PAI-AV OOD\\
          \midrule
          No reference & +4.54\% & +7.03\%\\
          Matched distractor & +3.15\% & +7.67\%\\
          Nearest entity & +1.43\% & +2.61\%\\
          Highest predicted risk & +1.26\% & +1.51\%\\
          \bottomrule
        \end{tabular*}
      \end{minipage}\hfill
      \begin{minipage}[t]{0.45\linewidth}
      \renewcommand{\arraystretch}{1.3}
        \begin{tabular*}{\linewidth}{@{\extracolsep{\fill}}lc@{}}
          \toprule
          \rowcolor{dsnormal}
          \multicolumn{2}{l}{\textbf{(b) Selective Correction}}\\
          Outcome & Rate\\
          \midrule
          Safe $\rightarrow$ Unsafe (Collision or Off-road) & 2.2\%\\
          Preserved (0m; exactly the same) & 53.9\%\\
          Preserved (3m) & 70.4\%\\
          \bottomrule
        \end{tabular*}
      \end{minipage}
      \end{minipage}
      \par\smallskip\raggedright
      \par
      \endgroup
    \end{minipage}
\end{figure}
\FloatBarrier

\begin{figure}[h]
  \centering
  \includegraphics[width=.9\linewidth]{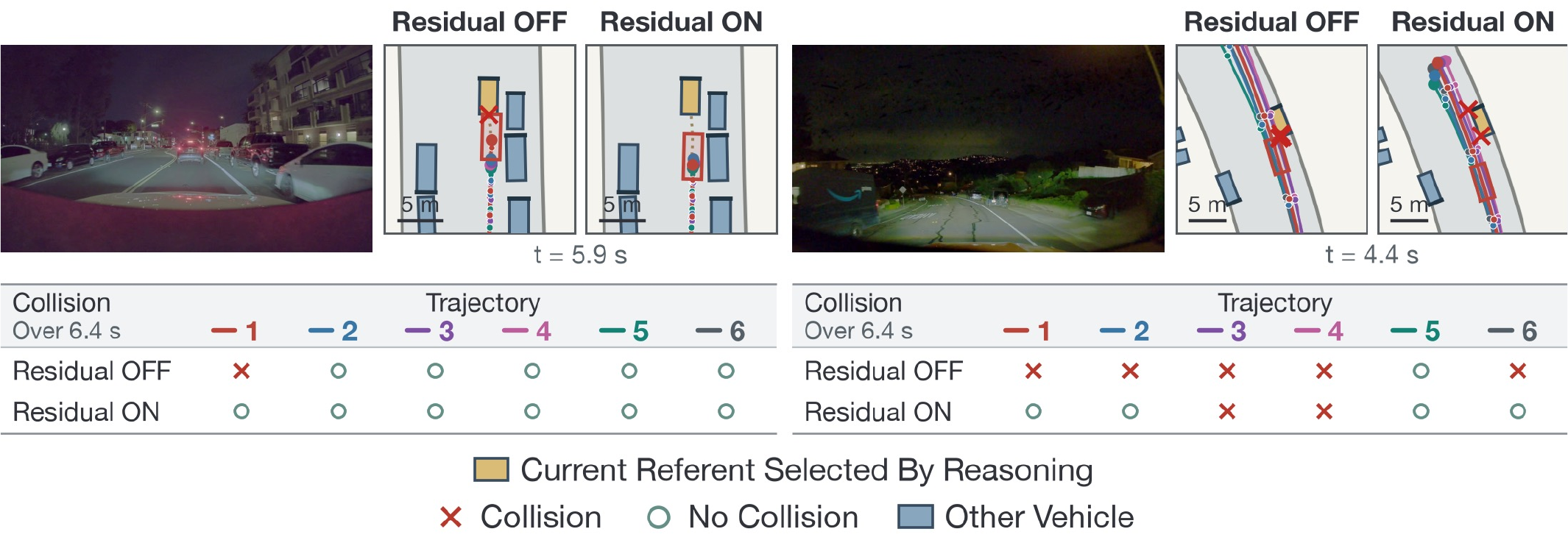}
  \caption{\textbf{Selective Correction.}
  BEV details show six matched trajectories.}
  \label{fig:selectivity-reference-analysis}
\end{figure}

\vspace{-4pt}
\paragraph{Selective Correction and Preservation.}
The final policy can repair unsafe plans while preserving some already-safe outputs, but may also introduce new safety issues (Table~\ref{tab:reference-selectivity}(b)) or leave existing ones insufficiently corrected (Figure~\ref{fig:selectivity-reference-analysis}).
Preservation reflects both correction and final arbitration and should therefore be attributed to the complete policy. We discuss these failures further in \S\ref{sec:eval-additional}.

\vspace{-4pt}
\paragraph{Grounding Reasoning through Entity References.}
Figure~\ref{fig:gounded_reasoning} illustrates how GroundAct references specific 
perceived entities when reasoning about following, lane keeping, lane changing, and stopping. 
These references enable retrieval of the corresponding continuous physical states, 
grounding subsequent reasoning and planning in the entities selected during reasoning. 
Their physical states and interactions then guide downstream decisions and shape the evolving plan.


\Needspace{14\baselineskip}
\subsection{Additional Analysis}
\label{sec:eval-additional}

\edef\evalanalysisbasewidth{\the\dimexpr0.49\linewidth\relax}
\edef\evalanalysistablewidth{\the\dimexpr0.42\linewidth\relax}
\edef\evalanalysistextwidth{\the\dimexpr\linewidth-\evalanalysistablewidth-12bp\relax}
\definecolor{evalaction}{HTML}{EAF3EF}
\definecolor{evalgt}{HTML}{F8F1E5}

\noindent\begin{minipage}{\linewidth}
  \begin{minipage}[t]{\evalanalysistextwidth}
    
    \paragraph{Reasoning Alignment with Predicted and GT Behaviors.}
    We further investigate how grounding affects reasoning--action alignment
    and alignment with ground-truth behavior. Specifically, we measure the
    consistency between meta-actions (e.g., stop, decelerate) inferred from
    the reasoning and those derived from either the predicted or ground-truth
    trajectories (Table~\ref{tab:eval-alignment}).
    Our method yields substantial improvements in both settings.
    We attribute these gains to two key factors:
    (1) improved spatial awareness of surrounding objects and
    (2) an explicit grounding chain that links reasoning to planning.
  \end{minipage}\hspace{12bp}%
  \begin{minipage}[t]{\evalanalysistablewidth}
    \vspace{-16pt}
    \centering
    \begingroup
      \expandafter\def\csname @captype\endcsname{table}

      \caption[Meta-action alignment.]{
        \raggedright
        \textbf{Reasoning-Action Alignment.}
        \evalmodeD{} discrete; \evalmodeC{} continuous.
         Appendix~\ref{app:additional-analysis}.
      }
      \label{tab:eval-alignment}

      \resizebox{\linewidth}{!}{%
      \begin{minipage}{\evalanalysisbasewidth}
        \centering
        \footnotesize
        \setlength{\tabcolsep}{3pt}
        \renewcommand{\arraystretch}{1.18}

        \begin{tabular*}{\linewidth}
          {@{\extracolsep{\fill}}lrrr@{}}
          \toprule
          Model & F1 & Recall & Precision \\
          \midrule

          \rowcolor{evalaction}
          \multicolumn{4}{l}{
            \textit{vs. meta-action from predicted trajectory}
          } \\
          Alpamayo-1.5~\evalmodeD{}
            & 55.98 & 47.24 & \textbf{68.70} \\
          Alpamayo-1.5~\evalmodeC{}
            & 51.61 & 44.74 & 60.96 \\
          \textbf{\method{} (Ours)}~\evalmodeC{}
            & \textbf{68.44} & \textbf{69.38} & 67.52 \\

          \midrule

          \rowcolor{evalgt}
          \multicolumn{4}{l}{
            \textit{vs. meta-action from GT trajectory}
          } \\
          Alpamayo-1.5~\evalmodeD{}
            & 58.48 & 48.77 & 73.03 \\
          Alpamayo-1.5~\evalmodeC{}
            & 58.52 & 49.06 & 72.51 \\
          \textbf{\method{} (Ours)}~\evalmodeC{}
            & \textbf{73.79} & \textbf{71.81} & \textbf{75.89} \\
          \bottomrule
        \end{tabular*}
      \end{minipage}%
      }
    \endgroup
  \end{minipage}
\end{minipage}\par

\noindent\begin{minipage}{\linewidth}
  \begin{minipage}[t]{\evalanalysistextwidth}
    \vspace{9pt}
    \paragraph{Object Detection and Referencing.}
    We follow Rex-Omni~\citep{rex-omni}, which evaluates object detection and pointing in 2D images, and adapt its evaluation protocol and metrics to BEV space.
    See Appendix~\ref{app:additional-analysis} for evaluation settings and baseline details.
    As shown in Table~\ref{tab:eval-perception}, \method{} performs strongly on both detection and geometry-aware referencing of planning-critical entities.
    These results indicate that our entity representation captures both semantic identity and spatial geometry, providing the physical information needed to ground prediction and planning in entity states and compute their physical risks.
    
  \end{minipage}\hspace{12bp}%
  \begin{minipage}[t]{\evalanalysistablewidth}
    \centering
    \begingroup
      \expandafter\def\csname @captype\endcsname{table}
      \caption[Detection and Referencing.]{\textbf{Detection and Referencing.}\protect\\
      Appendix~\ref{app:additional-analysis} for baseline details.}
      \label{tab:eval-perception}
      \resizebox{\linewidth}{!}{%
      \begin{minipage}{\evalanalysisbasewidth}
      \centering
      \footnotesize
      \setlength{\tabcolsep}{2pt}
      \renewcommand{\arraystretch}{1.05}
      \begin{tabular*}{\linewidth}{@{\extracolsep{\fill}}lrrrrr@{}}
        \toprule
        & \multicolumn{2}{c}{\cellcolor{evaldetect}\textbf{Detection (\%)}}
        & \multicolumn{3}{c}{\cellcolor{evalrefer}\textbf{Referencing (\%)}} \\
        \cmidrule(lr){2-3}\cmidrule(l){4-6}
        Method & mAP & \shortstack{NDS\\(BEV)}
        & \shortstack{F1@\\Point} & \shortstack{F1@\\IoU}
        & \shortstack{Critical\\mIoU} \\
        \midrule
        WildDet3D$^{z}$ & 11.71 & 17.07 & 28.00 & 19.43 & 52.99 \\
        FCOS3D & 27.07 & 44.40 & 49.30 & 36.25 & 69.22 \\
        MonoDETR & 21.98 & 39.43 & 43.26 & 30.72 & 74.52 \\
        PETR & 33.52 & 46.75 & 41.11 & 26.50 & 69.02 \\
        PETRv2 & 31.27 & 47.95 & \textbf{55.02} & 40.09 & 70.29 \\
        \midrule
        \textbf{\method{}} & \textbf{35.96} & \textbf{48.65} & 54.52
        & \textbf{42.44} & \textbf{75.20} \\
        \bottomrule
      \end{tabular*}
      \par\raggedright
      \vspace{4pt}
       $^{*}$ WildDet3D$^{z}$: zero-shot; others: fine-tuned.\par
      \end{minipage}%
      }%
    \endgroup
  \end{minipage}
\end{minipage}\par

\begin{figure}[t]
    \centering
    \includegraphics[width=1\linewidth]{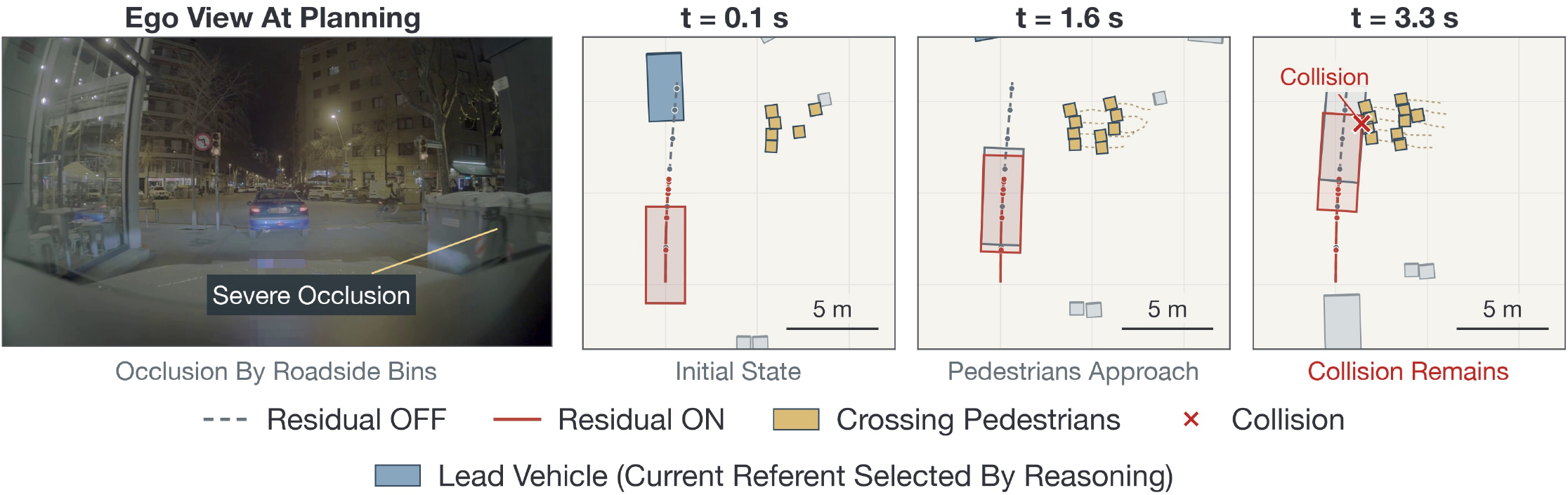}
    \vspace{-4pt}
    \caption{\textbf{Failure Case.} Pedestrians occluded at planning time emerge during execution.}
    \vspace{-8pt}
    \label{fig:failure_case}
\end{figure}

\vspace{4pt}
\textbf{Failure Case: Stale References.}
The correction acts only on entities referenced during reasoning, using their states at planning time. After the plan is generated, references can become outdated when occluded entities emerge (Figure~\ref{fig:failure_case}) or surrounding entities move rapidly. In either case, the correction may lack up-to-date referents to act on. Despite this limitation, GroundAct maintains strong planning performance in the safety-critical and OOD scenarios in Table~\ref{tab:eval-open-loop}. A natural next step is anticipatory reasoning: predicting future states to identify which entities will become planning-critical, and when, during execution.

%% file: sec/5_conclusion.tex
\vspace{-8pt}
\section{Conclusion}
\vspace{-8pt}
We revisit what groundedness should mean when the output is an action.
GroundAct preserves access to the continuous states of reasoning-selected
entities through Token-as-Reference, allowing their predicted interactions
to selectively shape the ego vehicle's evolving plan. Open- and closed-loop
evaluations show strong planning performance, while a controlled audit shows that 
disabling the referencing pathway from reasoning to planning can compromise driving safety.
These results support \emph{grounded planning}: preserving the physical
connection between what reasoning identifies as relevant and how the
vehicle acts.

%% file: sec/X_suppl.tex
\newpage
\appendix

\section{Grounding Criteria And Transfer Gap}
\label{app:transfer_gap}

An \emph{actor} is an annotated non-ego traffic participant with a GT track ID
(e.g., a vehicle, pedestrian, or cyclist. We manually designate
each scene's \emph{target actor} and constraint-relevant physical fact based on driving situation. Using each model's native
representation (Table~\ref{tab:grounding_criteria}), we assign $G=1$ for correct
grounding and $G=0$ for an established failure; missing or ambiguous results.

\begin{table}[ht]
  \centering
  \caption{\textbf{Model-Specific Grounding Criteria.}}
  \vspace{8pt}
  \label{tab:grounding_criteria}
  \small
  \setlength{\tabcolsep}{5pt}
  \renewcommand{\arraystretch}{1.1}
  \begin{tabularx}{\linewidth}{@{}lY@{}}
    \toprule
    Model & Criterion For Correct Grounding \\
    \midrule
    \begin{tabular}[t]{@{}l@{}}Alpamayo-1.5 \\ \&~SpanVLA\end{tabular} &
      Reasoning correctly identifies the target actor and relevant physical
      fact at the specified time, without contradiction. \\
    \addlinespace[3pt]
    FSDrive &
      The decoded future image at approximately $+0.5$\,s preserves actor
      class, coarse location relative to ego and road, and local road geometry. \\
    \addlinespace[3pt]
    GroundAct &
      Entity-reference validation requires a valid generated reference to the target actor in the same-run entity representation, a correct relevant fact, and BEV IoU $\geq 0.5$ with GT after temporal alignment.
      The above text criterion is used to validate textual reasoning.\\
    \bottomrule
  \end{tabularx}
\end{table}

For candidate $i$, $C_i=1$ denotes ego--actor footprint overlap or contact
within $0.1$--$5.0$\,s, evaluated at 10\,Hz without at-fault attribution.
Let $\mathcal{E}$ denote candidates with both grounding and collision labels. TransferGap is then computed as:
\[
  \mathrm{TransferGap}
  = \frac{\sum_{i\in\mathcal{E}} G_i C_i}
         {\sum_{i\in\mathcal{E}} G_i}.
\]

\section{Extended Related Work}
\label{app:extended-related-work}


\paragraph{Prediction and Interaction-Aware Planning.}
Future-aware driving methods use visual futures, latent dynamics, or
action-conditioned world models to inform planning
\citep{drivevla-w0,driveworld-vla}. UniAD and GameFormer
couple perception, multi-agent prediction, interaction, and planning through
query-based or game-theoretic structures~\citep{uniad,gameformer}, while P3 exposes
semantic occupancy as a differentiable planning cost~\citep{p3}. These works establish
that structured prediction and physical interaction can directly shape planning.
GroundAct asks a complementary question: \emph{which} physical entities should receive
explicit interaction-mediated computation, and how should that choice remain connected
to reasoning? We use reasoning-referenced entities as the interface to downstream
future and interaction modeling, making the selected grounding explicit in the
planning computation.

\paragraph{Reasoning-Guided Generative Planning.}
Flow matching and diffusion-based policies generate trajectories through iterative
refinement~\citep{flowmatching,scorediffusion}, while guidance and residual
pathways can modify a pretrained generator or controller without replacing its base
behavior~\citep{controlnet,cosmos-transfer1,tc-lora}. Streaming Intent further connects
reasoning to generative planning by deriving a discrete driving intent from
chain-of-thought reasoning and using classifier-free guidance to steer a rectified-flow
trajectory generator~\citep{actionintent}. GroundAct shares the goal of making
reasoning consequential for generative action, but differs in the mediator. Rather than
steering generation with a global semantic intent, GroundAct preserves references to
individual physical entities, evaluates their interactions with the evolving ego
proposal, and uses the resulting physical signal for selective residual correction.

\section{Training Implementation Details}
\label{app:training-recipe}

This section specifies the targets, loss reductions, and
optimization scope underlying Section~\ref{sec:training-objectives}. The
objectives preserve the physical correspondence as the representation,
reasoning interface, and planner are adapted.

\begingroup
\paragraph{Physical Labels and Losses.}
For the training objective, we follow the prior query-based object detection framework~\citep{detr}.

Let $\mathcal A$ denote the Hungarian assignment between entity slots and
annotated objects. Matched queries receive their objects' class and oriented
BEV box, and unmatched queries receive the null class. For class logit $z_{ic}$ and one-hot
target $y_{ic}$, including the null category, define
$p_{ic}=\sigma(z_{ic})$,
$p_{ic}^{\,t}=y_{ic}p_{ic}+(1-y_{ic})(1-p_{ic})$, and
$\alpha_{ic}^{\,t}=y_{ic}\alpha_f+(1-y_{ic})(1-\alpha_f)$.
The per-scene classification and box losses are
\begin{equation}
\begin{aligned}
\mathcal L_{\mathrm{cls}}
&=-\frac1K\sum_{i=1}^{K}\sum_{c=1}^{C+1}
\alpha_{ic}^{\,t}(1-p_{ic}^{\,t})^{\gamma_f}\log p_{ic}^{\,t},\qquad
\mathcal L_{\mathrm{box}}
&=\frac{1}{6\max(1,|\mathcal A|)}
\sum_{(i,j)\in\mathcal A}\|\hat b_i-b_j^\star\|_1.
\end{aligned}
\label{eq:train-physical-detail}
\end{equation}
Here $\alpha_f=0.25$ and $\gamma_f=2$ are the focal-loss parameters, and $\eta_{\mathrm{box}}=5$ is the box regression weight.
The box code contains normalized $(c_x,c_y,l,w,\sin\psi,\cos\psi)$.
Losses are averaged over scenes; matching combines class compatibility and box
distance.



\paragraph{Reference Targets and Reduction.}
Each annotated referent is matched to the class-consistent entity within a BEV-center distance threshold (2\,m). 
All $K=64$ entity states participate in
\Eqref{eq:train-association}. A reference without a valid physical match
is excluded from association and gate losses.

\paragraph{Association and Gating.}
Let $\mathcal V$ contain
reference positions with a valid match, and let $i_t^\star$ be the matched slot
for target token $y_{t+1}$. The association objective is
\begin{equation}
\mathcal L_{\mathrm{assoc}}
=-\frac{1}{\max(1,|\mathcal V|)}\sum_{t\in\mathcal V}
\log\frac{\exp[\tau\cos(W_qh_t,W_kM_{i_t^\star})]}
{\sum_{j=1}^{K}\exp[\tau\cos(W_qh_t,W_kM_j)]}.
\label{eq:train-association}
\end{equation}
Gate supervision balances reference positions $\mathcal G_1$ and ordinary-text
positions $\mathcal G_0$. With
$\mathcal C=\{b\in\{0,1\}:|\mathcal G_b|>0\}$, the objectives are
\begin{equation}
\begin{aligned}
\mathcal L_{\mathrm{gate}}
&=-\frac{1}{\max(1,|\mathcal C|)}\sum_{b\in\mathcal C}
\frac{1}{|\mathcal G_b|}\sum_{t\in\mathcal G_b}
\left[b\log\pi_t+(1-b)\log(1-\pi_t)\right],\\
\end{aligned}
\label{eq:train-reference-detail}
\end{equation}
 If no gate labels are valid, its loss is zero.
The index sets $\mathcal V$ and $\mathcal G_b$ collect valid
entries across the current training microbatch, with scene indices suppressed.
These independently normalized losses form the reference objective in \Eqref{eq:train-reasoning}.


\paragraph{Token Prediction.}
Ordinary language targets exclude reference-token positions. Let
$\mathcal T_{\mathrm{text}}$ and $\mathcal T_{\mathrm{action}}$ denote the
supervised non-reference text and discrete-action positions. Their masked
next-token losses are reduced separately:
\begin{equation}
\mathcal L_{\mathrm{text}}
=\sum_{q\in\{\mathrm{text},\mathrm{action}\}}
\eta_q
\left[-\frac{1}{\max(1,|\mathcal T_q|)}
\sum_{t\in\mathcal T_q}
\log p_{\mathrm{LM},t}(y_{t+1}^\star)\right].
\label{eq:train-text-detail}
\end{equation}
Here $p_{\mathrm{LM},t}$ is the vocabulary-head softmax over the full token
vocabulary, and $\mathcal T_q$ collects supervised positions across the current
training microbatch. Reference supervision is excluded through target-position
masking. The prefill objective $\mathcal L_{\mathrm{LM}}$ uses the same weighted,
separately reduced cross-entropy form on the original supervised token groups,
before reference adaptation.
The balanced gate loss averages the positive and negative group means when
both are present and uses the remaining mean when only one group is present.

\paragraph{Initializing Address Feedback.}
Association alignment first learns the pointer projections against a fixed
entity representation. The feedback projection is additionally initialized
to make a consumed reference resolve to its own entity. Let
$a_i=b_i+P(\operatorname{sg}[M_i])$ be the address representation before feedback
interpolation. With the pointer fixed and gradients stopped through the entity
bank, address retrieval uses
\begin{equation}
\mathcal L_{\mathrm{addr}}
=-\frac{1}{\max(1,|\mathcal V_{\mathrm{addr}}|)}
\sum_{i\in\mathcal V_{\mathrm{addr}}}
\log p_{\mathrm{ptr}}^{\mathrm{fixed}}
\!\left(i\mid a_i,\operatorname{sg}[\mathcal M]\right).
\label{eq:train-address-detail}
\end{equation}
This objective trains the input projection to preserve the intended address.
Subsequent teacher-forced reasoning consumes the corresponding live entity
state at the next token position, matching the reference-consumption operation
used during generation.

\paragraph{Base Action Expert Training.}
The expert learns recorded actions $a^\star$ through conditional flow matching,
conditioned on the complete frozen VLM key/value context $c$, including
observations and reasoning:
\begin{equation}
x_s=(1-s)\epsilon+s a^\star,\qquad
\mathcal L_{\mathrm{AE}}
=\mathbb E_{a^\star,c,\epsilon,s}
\left\|v_\omega(x_s,s;c)-(a^\star-\epsilon)\right\|_{\mathrm{av}}^2.
\label{eq:train-ae}
\end{equation}
Here $\epsilon\sim\mathcal N(0,I)$, $s$ is noise-to-action flow time, and the
squared norm averages over action coordinates and horizon positions.
The objective is evaluated in the action expert's encoded, normalized action
space; geometric checks operate on decoded ego trajectories.
The expectation samples Gaussian noise and flow time $s\sim p_s$.
Flow time is distinct from physical future time and pointer sharpness $\tau$.
The stored flow-matching sampler uses a transformed beta distribution,
$s=s_{\max}(1-B)$ with $B\sim\operatorname{Beta}(a_s,b_s)$, rather than a
uniform flow-time distribution. Its source configuration sets
$s_{\max}=0.999$, $a_s=1.5$, and $b_s=1$.
The trained expert remains frozen during physical prediction and residual learning.

\paragraph{Physical-Prediction Supervision.}
For the forecast-based formulation described in \S\ref{sec:act}, let
$S_{ih}^\star$ be the annotated state of the object matched to slot $i$ at
physical horizon index $h$, and let $m_{ih}$ mark valid supervision.
The forecast objective can be written as
\begin{equation}
\mathcal L_{\mathrm{future}}
=\mathbb E\left[
\frac{\sum_{i,h}m_{ih}\,
\ell_{\mathrm{state}}(\hat S_{ih}^{(k)},S_{ih}^\star)}
{\max(1,\sum_{i,h}m_{ih})}\right].
\label{eq:train-future}
\end{equation}
The state penalty supervises the physical quantities represented by the forecast.
The expectation includes training examples and sampled proposal states indexed
by $k$; $h$ indexes physical time, whereas $k$ indexes refinement of the proposal.

\paragraph{Teacher Construction and Training Inputs.}
Teachers are selected from frozen-planner candidates, and selection uses a $4$\,s
evaluation horizon.
The recorded teacher selector requires finite actions and trajectories,
action bounds, a target-clearance gain of at least $0.5$\,m, and a progress ratio
of at least $0.95$ relative to the base proposal. If the base overlaps the
target entity, the alternative must remove that overlap. It must also introduce
no newly overlapping entity identities across the scene. Each stored teacher
record retains the flow state, base field, flow time, integration step, and
physical features used to construct its target.

At a stored state $(x_s,s)$, let $u_s^\star$ denote
the teacher field and define its raw and applied residual targets as
\begin{equation}
r_s^\star=\operatorname{sg}\!\left[
u_s^\star-v_{\mathrm{base}}(x_s,s;c)\right],
\qquad d_s^\star=\gamma_s^\star r_s^\star.
\label{eq:train-teacher}
\end{equation}
The base and teacher fields are evaluated at the same state in the same action
coordinates. The teacher gate $\gamma_s^\star$ determines how much of the raw
correction is applied.

For a retained alternative $a^+$, the recorded endpoint-based field and gate are
\begin{equation}
u_s^\star=\frac{a^+-x_s}{\max(1-s,\delta)},\qquad
\gamma_s^\star=b_s^{\mathrm{risk}}\min(1,\kappa s^2),\qquad
d_s^\star=\gamma_s^\star
\operatorname{sg}[u_s^\star-v_{\mathrm{base}}].
\label{eq:train-teacher-detail}
\end{equation}
The binary risk latch $b_s^{\mathrm{risk}}$ remains active once a teacher trace
has triggered correction. The recorded ten-step configuration uses
$\delta=0.1$ and $\kappa=2$. The denominator floor limits endpoint conversion
near the end of the flow.

\paragraph{Residual Distillation and Risk-Off Preservation.}
Let $z_{\mathcal R}^{\mathrm{train}}$ denote the ground-truth-derived interaction
features described above. With adapter prediction
$\hat r_s=R_\phi(x_s,v_{\mathrm{base}},z_{\mathcal R}^{\mathrm{train}})$ and
$\hat d_s=\gamma_s^\star\hat r_s$, distillation uses
\begin{equation}
\mathcal L_{\mathrm{field}}
=\mathbb E_{\mathcal D_+}\|\hat d_s-d_s^\star\|_{\mathrm{av}}^2,
\qquad
\mathcal L_{\mathrm{raw}}
=\mathbb E_{\mathcal D_{\mathrm{on}}}\|\hat r_s-r_s^\star\|_{\mathrm{av}}^2.
\label{eq:train-correction}
\end{equation}
Here $\mathcal D_+$ contains retained teacher traces and
$\mathcal D_{\mathrm{on}}$ their risk-active states. Applying the teacher gate
to both sides preserves the target's intended correction strength; the raw
term additionally supervises the adapter before gating.

For risk-inactive states $\mathcal D_{\mathrm{off}}$ in the teacher traces,
preservation discourages a nonzero raw correction:
\begin{equation}
\mathcal L_{\mathrm{off}}
=\mathbb E_{\mathcal D_{\mathrm{off}}}\|\hat r_s\|_{\mathrm{av}}^2.
\label{eq:train-risk-off-detail}
\end{equation}
Each masked term is normalized by the number of selected scalar elements in
the minibatch and is zero when none are selected. These penalties supervise the correction field.

\paragraph{Residual Consistency and Preservation.}
Let $\Delta s$ be the integration increment recorded with a teacher state.
The predicted and teacher next states are
$\hat x'=x_s+\Delta s(v_{\mathrm{base}}+\hat d_s)$ and
$x^{\star\prime}=x_s+\Delta s(v_{\mathrm{base}}+d_s^\star)$.
With $D_h^2$ the second difference along the physical action horizon, the
remaining terms in \Eqref{eq:train-preservation} are
\begin{equation}
\begin{aligned}
\mathcal L_{\mathrm{step}}&=\mathbb E\|\hat x'-x^{\star\prime}\|_{\mathrm{av}}^2,\\
\mathcal L_{\mathrm{smooth}}&=\mathbb E\|D_h^2\hat d_s-D_h^2d_s^\star\|_{\mathrm{av}}^2,
\qquad
\mathcal L_{\mathrm{norm}}=\mathbb E\|\hat r_s\|_{\mathrm{av}}^2,\\
\mathcal L_{\mathrm{reg}}
&=\lambda_{\mathrm{step}}\mathcal L_{\mathrm{step}}
+\lambda_{\mathrm{smooth}}\mathcal L_{\mathrm{smooth}}
+\lambda_{\mathrm{norm}}\mathcal L_{\mathrm{norm}}.
\end{aligned}
\label{eq:train-regularization-detail}
\end{equation}
The recorded distillation configuration uses
\begin{equation}
(\lambda_{\mathrm{field}},\lambda_{\mathrm{raw}},
\lambda_{\mathrm{step}},\lambda_{\mathrm{off}},
\lambda_{\mathrm{smooth}},\lambda_{\mathrm{norm}})
=(1,\;0.25,\;2,\;0.1,\;0.01,\;0.001).
\label{eq:train-distillation-weights}
\end{equation}
The risk-off penalty acts on the raw adapter output, so it remains informative
when the applied correction is zeroed by the gate. It encourages preservation
without asserting an explicit loss on the final integrated action.

\begin{table}[htbp]
  \centering
  \caption{\textbf{Optimization Scope.}
  Each interface is adapted while preserving the upstream information it uses.}
  \label{tab:act-optimization-scope}
  \small
  \setlength{\tabcolsep}{4pt}
  \renewcommand{\arraystretch}{1.18}
  \begin{tabularx}{\linewidth}{@{}l X X@{}}
    \toprule
    Interval & Trainable parameters & Fixed components \\
    \midrule
    Entity warm-up & Entity queries and class/box heads & Vision and language backbones \\
    Joint adaptation & Backbones, queries, physical heads, and vocabulary head & --- \\
    Association alignment & Pointer projections and temperature & Entity encoder and language model \\
    Language adaptation & Language decoder, vocabulary head, and reference gate & Vision encoder, queries, physical heads, and pointer parameters \\
    Address initialization & Feedback projection & Pointer and source entity states \\
    Action learning & Base action expert & Multimodal encoder and reasoner \\
    Physical prediction & Prediction module & Base planner \\
    Residual distillation & Residual adapter & Base planner and teacher records \\
    \bottomrule
  \end{tabularx}
\end{table}

\endgroup


\section{Experimental Setup Details}
\label{app:experimental-setup-details}

\paragraph{Baseline Configurations.}
Alpamayo-R1, Alpamayo 1.5, Alpamayo 2 Super, and LCDrive are evaluated from their
authors' released checkpoints without fine-tuning; all were trained on PAI-AV, with
training sets that differ in size from ours. The evaluated LCDrive uses Qwen2.5-0.5B
and DINOv2-Small, with approximately 0.6B parameters. FSDrive$^{\dagger}$, AutoVLA$^{\dagger}$ and SpanVLA$^{\dagger}$ share \method{}'s Qwen3-VL-2B backbone: we re-implement each method's defining
component on this backbone (e.g., action tokenization) and fine-tune the resulting model on the same 140.5K PAI-AV training samples as \method{}, without subsequent reinforcement learning.

\paragraph{Open-Loop Sample Selection.}
Normal evaluation uses the first 4,000 samples of the PAI-AV default split, and OOD
evaluation uses its 290-sample OOD split. The PAI-AV$^{*}$ set contains 512 high-risk
samples (Appendix~\ref{app:safety-critical-distribution}).

\paragraph{Computational Setup.}
Training uses 32 NVIDIA A100-SXM4-80GB GPUs throughout all stages.
For each evaluated sample, open-loop evaluation runs on a single
NVIDIA A100-SXM4-80GB GPU and closed-loop evaluation on a single
NVIDIA H100 80GB node.

\section{Safety-Critical Evaluation Dataset}
\label{app:safety-critical-distribution}

\begin{figure}[h]
  \centering
  \includegraphics[width=0.7\linewidth]{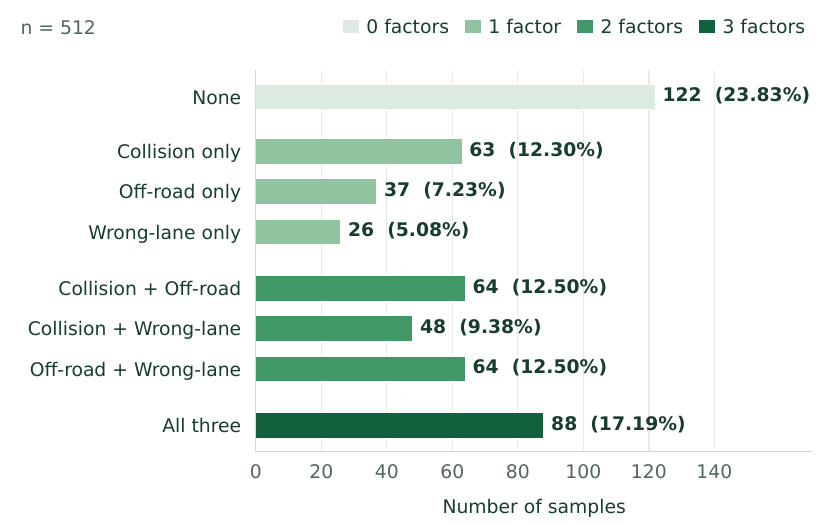}
  \caption{\textbf{Distribution of safety-critical factors in PAI-AV$^{*}$ ($n=512$).}
    Factors are identified from the open-loop plans generated by the released Alpamayo-R1 10B model.
    Bars represent mutually exclusive factor combinations, with labels indicating counts and percentages and colors indicating the number of factors.
    ``None’’ denotes samples with none of the listed factors.}
  \label{fig:eval-safety-critical-distribution}
\end{figure}

The PAI-AV$^{*}$ open-loop evaluation set contains 512 PAI-AV samples selected as
high-risk scenes; none overlaps the 140.5K
training samples used by \method{}, FSDrive$^{\dagger}$, AutoVLA$^{\dagger}$, and SpanVLA$^{\dagger}$.
Figure~\ref{fig:eval-safety-critical-distribution} characterizes the set through the
open-loop plans of the released Alpamayo-R1 10B checkpoint: at least one of collision,
off-road, or wrong-lane occurs in 390 samples (76.17\%), including 264 (51.56\%) with
two or three factors, while 122 (23.83\%) show none. These factors describe a baseline's
behavior on the set rather than a property of the scenes.

\section{Reference-Objective Evaluation Metrics}
\label{app:reference-objective-metrics}


\begin{table}[ht]
  \centering
  \caption{\textbf{Reference-Objective Evaluation Metrics.}
  All five reported scores are percentages.
  $TP$, $FP$, and $FN$ denote matched predictions, unmatched predictions,
  and missed targets in physical-reference matching.}
  \vspace{8pt}
  \label{tab:reference-objective-metric-definitions}
  \small
  \setlength{\tabcolsep}{4pt}
  \renewcommand{\arraystretch}{1.5}
  \begin{tabular}{@{}p{0.23\linewidth}p{0.71\linewidth}@{}}
    \toprule
    Metric & Definition \\
    \midrule
    Valid Emission &
    Validity of the references emitted in generated reasoning. \\
    Sentence Complete &
    Completion of generated reasoning, assessed separately from the
    correctness of its physical references. \\
    Grounding Precision &
    Count-based precision:
    $100\,TP/(TP+FP)$, where $TP$ and $FP$ denote matched and unmatched
    predicted physical referents. \\
    Grounding Recall &
    Count-based recall:
    $100\,TP/(TP+FN)$, where $FN$ denotes unmatched target referents. \\
    Grounding F1 &
    Count-based harmonic mean:
    $100\,2TP/(2TP+FP+FN)$. \\
    \bottomrule
  \end{tabular}
\end{table}



\section{Metrics for Open-Loop Planning Evaluation}
Each sample is a clip and a prediction timestamp. Every model generates six
candidate trajectories per sample, each with 64 poses at 10\,Hz (6.4\,s).
Table~\ref{tab:open-loop-metric-definitions} defines the four metrics. Each
metric is averaged over the six candidates of a sample and then over samples. 
Collision and off-road are per-candidate binary
events, counted once per candidate, evaluated at the discrete poses without
at-fault filtering, and reported as percentages. ADE and corner distance use
all samples; collision uses the annotation-available subsets.

\begin{table}[h]
  \centering
  \caption{\textbf{Open-loop evaluation metrics.} Lower is better for all four.}
  \vspace{4pt}
  \label{tab:open-loop-metric-definitions}
  \small
  \begin{tabularx}{\linewidth}{@{}l l X l@{}}
    \toprule
    \textbf{Metric} & \textbf{Horizons} & \textbf{Definition} & \textbf{Samples} \\
    \midrule
    \textbf{ADE (m)} $\downarrow$ & 2.5, 5.0, 6.4\,s &
      Mean ground-plane distance between predicted and ground-truth ego positions over the horizon. &
      All \\
    \textbf{Corner distance (m)} $\downarrow$ & 6.4\,s &
      The same distance averaged over the four corners of the oriented ego box. &
      All \\
    \textbf{Collision (\%)} $\downarrow$ & 2.5, 5.0\,s &
      Fraction of candidates whose predicted ego footprint intersects any annotated actor footprint within the horizon. &
      3,846 / 267 / 512\textsuperscript{a} \\
    \textbf{Off-road (\%)} $\downarrow$ & 2.5, 5.0\,s &
      Fraction of candidates whose predicted ego footprint contains a road-boundary point within the horizon. &
      PAI-AV$^{*}$ only\textsuperscript{b} \\
    \bottomrule
  \end{tabularx}
  \par\smallskip
  {\footnotesize \textsuperscript{a}Annotation-available subsets of Normal / OOD / PAI-AV$^{*}$.
   \textsuperscript{b}Normal and OOD lack map data.}
\end{table}

\section{Metrics for Closed-Loop Driving Evaluation}
\label{app:closed-loop-metrics}

For Table~\ref{tab:eval-closed-loop}, we evaluate each model on the same
$N=913$ \texttt{public\_2601} scenes in AlpaSim~\citep{alpasim}, with one rollout
per scene. The valid evaluation interval extends from the start of policy control
through the first collision of any type, off-road event, or deviation of at least
$4\,\mathrm{m}$ from the recorded ground-truth path. 
All metrics are computed using AlpaSim’s default protocol.

\begin{table}[H]
  \centering
  \small
  \caption{\textbf{Closed-loop evaluation metrics.} The two distance metrics are
  diagnostic and have no preferred direction.}
  \vspace{6pt}
  \label{tab:closed-loop-metric-definitions}
  \setlength{\tabcolsep}{4pt}
  \renewcommand{\arraystretch}{1.5}
  \begin{tabularx}{\linewidth}{@{}>{\raggedright\arraybackslash}p{0.27\linewidth}X@{}}
    \toprule
    \textbf{Metric} & \textbf{Definition} \\
    \midrule
    \textbf{Progress $\uparrow$} & Distance along the ground-truth path reached by projecting the final ego position onto the path, normalized by its total length. \\
    \textbf{At-fault collision $\downarrow$} & Fraction of scenes with at least one front or lateral ego collision. \\
    \textbf{Off-road $\downarrow$} & Fraction of scenes where the ego footprint contacts a road edge while not fully contained within the lane area. \\
    \textbf{Wrong lane $\downarrow$} & Fraction of scenes where the ego heading differs by more than $120^\circ$ from the direction of a lane containing the ego center. \\
    \textbf{Maximum Distance to GT Trajectory (m)} & Maximum distance from the ego center to the nearest point on the GT path. \\
    \textbf{Min. obstacle distance (m)} & Minimum distance between the ego footprint and any other actor's footprint. \\
    \bottomrule
  \end{tabularx}
\end{table}

\section{Setting Details of Additional Analysis}
\label{app:additional-analysis}

\paragraph{Perception Baselines.}
We fine-tune FCOS3D~\citep{fcos3d}, MonoDETR~\citep{monodetr},
PETR~\citep{petr}, and PETRv2~\citep{petrv2} from released checkpoints on
140,554 training frames, using one current front-wide image and camera
calibration. PETRv2 uses no temporal inputs. WildDet3D~\citep{wilddet3d}
is evaluated zero-shot. Table~\ref{tab:perception-training-settings}
summarizes training settings; model names link to official implementations.

\begin{table}[ht]
  \centering
  \caption{\textbf{Perception Baseline Training Settings.}
  Batch is global; LR and WD denote base learning rate and default weight decay.
  WildDet3D is evaluated without fine-tuning.}
  \label{tab:perception-training-settings}
  \small
  \begin{tabular*}{\linewidth}{@{\extracolsep{\fill}}llrrr@{}}
    \toprule
    Model / Backbone & Optimizer & Batch & LR & WD \\
    \midrule
    \href{https://github.com/open-mmlab/mmdetection3d}{FCOS3D} / R101-DCN
      & SGD & 16 & $2\!\times\!10^{-3}$ & $10^{-4}$ \\
    \href{https://github.com/ZrrSkywalker/MonoDETR}{MonoDETR} / R50
      & AdamW & 16 & $2\!\times\!10^{-4}$ & $10^{-4}$ \\
    \href{https://github.com/megvii-research/PETR}{PETR} / R50-DCN
      & AdamW & 8 & $2\!\times\!10^{-4}$ & $10^{-2}$ \\
    \href{https://github.com/megvii-research/PETR}{PETRv2} / V-99-eSE
      & AdamW & 8 & $2\!\times\!10^{-4}$ & $10^{-2}$ \\
    \href{https://github.com/allenai/WildDet3D}{WildDet3D}
      & -- & -- & -- & -- \\
    \bottomrule
  \end{tabular*}
\end{table}
\FloatBarrier

\paragraph{Detection And Referencing Metrics.}
We adapt nuScenes detection metrics~\citep{nuscenes} and Rex-Omni
pointing and overlap criteria~\citep{rex-omni} to BEV space
(Table~\ref{tab:perception-metric-definitions}). 
Tables~\ref{tab:eval-perception} and~\ref{tab:eval-alignment} report
percentages (higher is better), with the best value per metric and
evaluation target in bold.

\begin{table}[ht]
  \centering
  \caption{\textbf{Detection And Referencing Metrics.}
  All scores are reported as percentages. For the two F1 metrics,
  $F(H)=2H/(N_P+N_G)$, where $N_P$ and $N_G$ count retained predictions
  and GT objects, and $H$ counts GT objects hit at least once.}
  \vspace{8pt}
  \label{tab:perception-metric-definitions}
  \small
  \setlength{\tabcolsep}{5pt}
  \renewcommand{\arraystretch}{1.1}
  \begin{tabularx}{\linewidth}{@{}lY@{}}
    \toprule
    Metric & Definition \\
    \midrule
    mAP &
      Average precision using BEV center-distance matching, averaged
      over evaluated categories and thresholds $\{0.5,1,2,4\}$\,m. \\
    \addlinespace[3pt]
    NDS (BEV) &
      Combines mAP with BEV translation, scale, and orientation
      errors; velocity and attribute terms are omitted:
      $[5\,\mathrm{mAP}+
      \sum_{e\in\{\mathrm{mATE},\mathrm{mASE},\mathrm{mAOE}\}}
      \max(0,1-e)]/8$, using unscaled mAP. \\
    \addlinespace[3pt]
    F1@Point &
      $F(H)$ with a GT hit when any predicted center falls inside
      its BEV footprint, irrespective of category. \\
    \addlinespace[3pt]
    F1@IoU &
      $F(H)$ with a GT hit when any prediction has rotated BEV IoU
      $\geq 0.5$ with it, irrespective of category. \\
    \addlinespace[3pt]
    Critical mIoU &
      Mean BEV IoU for matched GT-CoC-referenced objects,
      excluding unmatched objects. Pairs are selected after class-agnostic,
      one-to-one center matching to all GT objects within 2\,m. \\
    \bottomrule
  \end{tabularx}
\end{table}
\FloatBarrier

\section{Inference Efficiency.}
\begingroup
\setlength{\columnsep}{10bp}
\setlength{\intextsep}{0pt}

\begin{wraptable}{r}{0.4\linewidth}
\centering

\fontsize{7.5}{9.0}\selectfont
\setlength{\tabcolsep}{1.5pt}
\setlength{\abovecaptionskip}{0pt}

\caption{\textbf{E2E Latency.}}
\label{tab:runtime_mini}
\label{tab:eval-runtime}
\vspace{2pt}

\begin{tabular}{@{}lcccr@{}}
\toprule
\textbf{Config.} & \textbf{Ground} & \textbf{Ref.} & \textbf{Act} & \textbf{Time (s)} \\
\midrule
\shortstack[l]{Driving VLM\\backbone}
& -- & -- & -- & $0.88 \pm 0.14$ \\
\midrule

\shortstack[l]{Spatial\\grounding}
& $\checkmark$ & -- & -- & $0.94 \pm 0.12$ \\
\midrule

\shortstack[l]{Grounded\\reasoning}
& $\checkmark$ & $\checkmark$ & -- & $0.99 \pm 0.14$ \\
\midrule
\textbf{GroundAct}
& $\checkmark$ & $\checkmark$ & $\checkmark$
& $\mathbf{1.10 \pm 0.19}$ \\
\bottomrule
\label{eval:latency}
\end{tabular}

\end{wraptable}

As shown in Table~\ref{eval:latency}, GroundAct adds $0.22$\,s to the backbone VLM's mean inference latency ($0.88$\,s), bringing the total to $1.10$\,s on a single NVIDIA A100-SXM4-80GB GPU without model- or inference-level optimization.
Potential avenues for reducing latency include compiled inference and TensorRT deployment. Beyond latency reduction, another
deployment direction is to decouple high-level planning from low-level control. GroundAct's strong open-loop planning performance over a $6.4$\,s prediction horizon motivates exploring lower-frequency grounded planning (e.g., approximately $1$\,Hz) paired with a faster low-level controller.
\par
\endgroup